\documentclass[10pt,twocolumn,letterpaper]{article}

\usepackage{cvpr}
\usepackage{times}
\usepackage{epsfig}
\usepackage{graphicx}
\usepackage{amsmath}
\usepackage{amssymb}

\usepackage{caption}
\usepackage{subcaption}
\usepackage{paralist}
\usepackage{array}
\usepackage{placeins}
\usepackage{multirow}
\usepackage{epstopdf}
\usepackage{paralist}
\usepackage[pagebackref=true,breaklinks=true,letterpaper=true,colorlinks,bookmarks=false]{hyperref}

\usepackage[ruled]{algorithm2e}
\makeatletter
\def\BState{\State\hskip-\ALG@thistlm}
\makeatother
\usepackage[pagebackref=true,breaklinks=true,letterpaper=true,colorlinks,bookmarks=false]{hyperref}

\cvprfinalcopy

\ifcvprfinal\fi
\usepackage{color}

\begin{document}

\title{LocNet: Improving Localization Accuracy for Object Detection}

\author{Spyros Gidaris\\
Universite Paris Est, Ecole des Ponts ParisTech\\
{\tt\small gidariss@imagine.enpc.fr}
\and
Nikos Komodakis\\
Universite Paris Est, Ecole des Ponts ParisTech\\
{\tt\small nikos.komodakis@enpc.fr}\\
}

\maketitle
\thispagestyle{empty}

\begin{abstract}
We propose a novel object localization methodology with the purpose of boosting the localization accuracy of state-of-the-art object detection systems.
Our model, given a search region, aims at returning the bounding box of an object of interest inside this region. 
To accomplish its goal, it relies on assigning conditional probabilities to each row and column of this region, where these probabilities provide useful
information regarding the location of the boundaries of the object inside the search region and  allow the accurate inference of the object bounding box under a simple probabilistic framework. 

For implementing our localization model, we make use of a convolutional neural network architecture that is properly adapted for this task, called LocNet. 
We show experimentally that LocNet achieves a very significant improvement on the mAP for high IoU thresholds on PASCAL VOC2007 test set and that it can be very easily coupled with recent state-of-the-art object detection systems, helping them to boost their performance. 
Finally, we demonstrate that our detection approach can achieve high detection accuracy even when it is given as input a set of {sliding windows}, thus proving that it is independent of box proposal methods.
\end{abstract}

\section{Introduction}
\begin{figure}[h]
\center
\renewcommand{\figurename}{Figure}
\renewcommand{\captionlabelfont}{\bf}
\renewcommand{\captionfont}{\small} 
\begin{center}
        \begin{center}
        \includegraphics[width=0.45\textwidth]{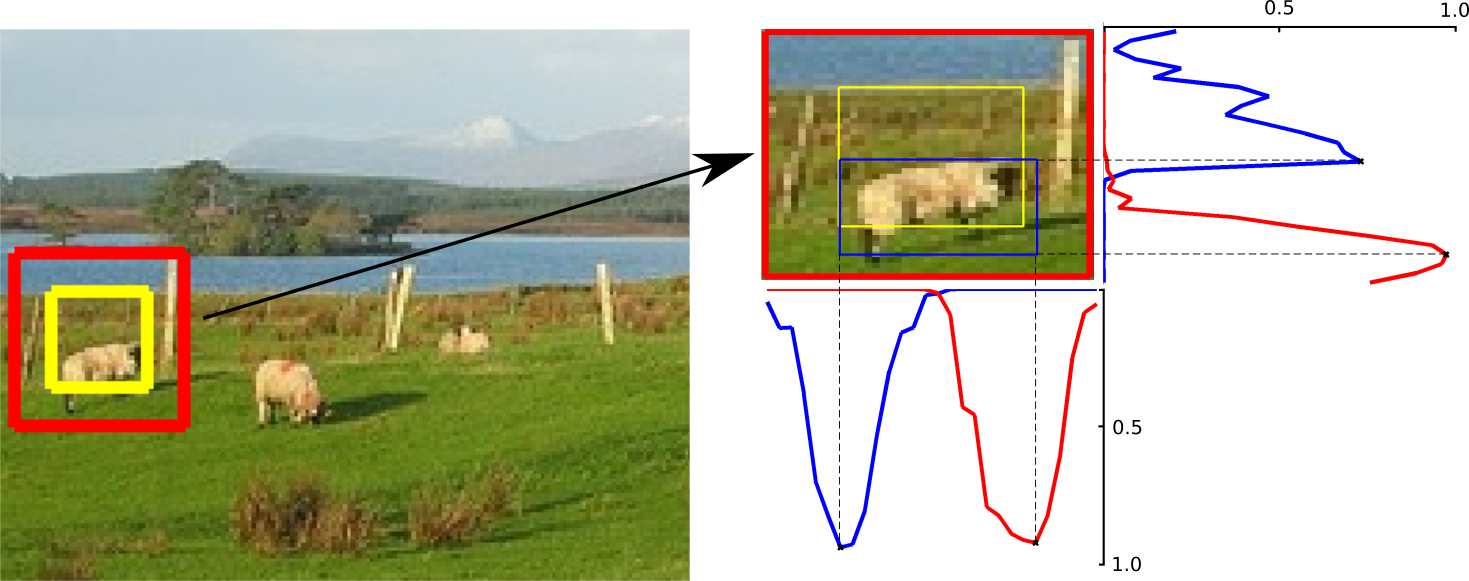}   
        \end{center}
        \begin{center}
        \includegraphics[width=0.45\textwidth]{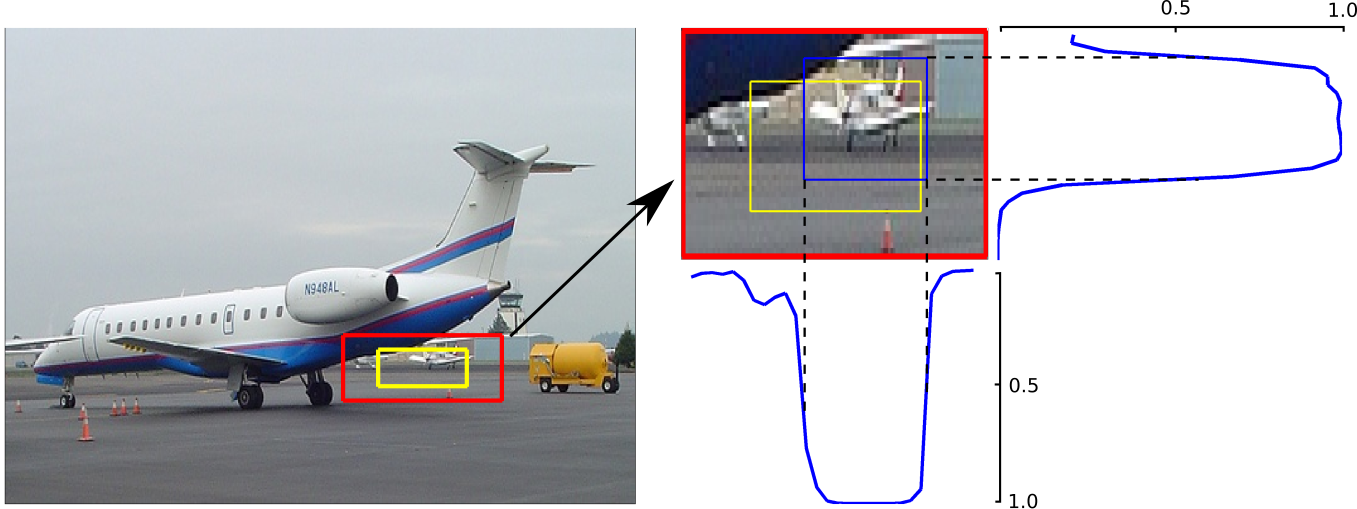}   
        \end{center}    
\end{center}             
\caption{Illustration of the basic work-flow of our localization module. 
\emph{\textbf{Left column:}} our model 
given a candidate box $B$ (yellow box) it "looks" on a search region $R$ (red box), which is obtained by enlarging box $B$ by a constant factor, in order to localize the bounding box of an object of interest. \emph{\textbf{Right column:}} To localize a bounding box the model assigns one or more probabilities on each row and independently on each column of region $R$. Those probabilities can be either
the probability of an element (row or column) to be one of the four object borders (see top-right image), or the probability for being on the inside of an objects bounding box (see bottom-right image). In either case the  predicted bounding box is drawn with blue color.}
\label{fig:LocModelInOut}        
\end{figure}

{\let\thefootnote\relax\footnotetext{This work was supported by the ANR SEMAPOLIS project. Code and trained models are available on:\\\url{https://github.com/gidariss/LocNet}}}

Object detection  is a computer vision problem that has attracted an immense amount of attention over the last years. The localization accuracy by which a detection system is able to predict the bounding  boxes of the objects of interest is typically judged based on the Intersection over Union  (IoU) between the predicted and the ground truth bounding box. 
Although in challenges such as  PASCAL VOC an IoU detection threshold of $0.5$ is  used  for deciding whether an object has been successfully detected, in real life applications a higher localization accuracy (e.g. IoU $\geqslant 0.7$) is  normally required (\eg, consider the task of a robotic arm that must grasp an object).
Such a need is also reflected in the very recently introduced \emph{COCO} detection challenge~\cite{lin2014microsoft}, which uses as evaluation metric the traditional average precision (AP) measurement but averaged over multiple IoU thresholds between 0.5 (loosely localized object) and 1.0 (perfectly localized object) so as to reward detectors that exhibit good localization accuracy.  

Therefore, proposing detectors that exhibit highly accurate (and not loose) localization of the ground truth objects should be one of the major future challenges in object detection. The aim of this work is to take a further step towards addressing this challenge. In practical terms,    our goal is to  boost   the bounding box detection AP  performance  across a wide range of IoU thresholds (\ie, not just for IoU threshold of $0.5$ but also for values well above that).
To that end, a main technical contribution of this work is to propose a novel \emph{object localization model} that, given a loosely localized search region inside an image, aims to return the accurate location of an object in this region (see Figure~\ref{fig:LocModelInOut}). 

A crucial component of this new model is that it does not rely on the commonly used  bounding box regression paradigm, which uses a regression function to directly predict the object bounding box coordinates.
Indeed, the motivation behind our work stems from the belief that trying to directly regress to the target bounding box coordinates, constitutes a  difficult learning task that cannot yield accurate enough bounding boxes. 
We argue that it is far more effective to attempt to localize a bounding box by first 
assigning a probability to each row and independently to each column of the search region  for being the left, right, top, or bottom borders of the bounding box (see Fig.~\ref{fig:LocModelInOut} top) or for being on the inside of an object's bounding box (see Fig.~\ref{fig:LocModelInOut} bottom). 
In addition, this type of probabilities  can provide a measure of confidence for placing the bounding box on each location and  they can also handle instances that exhibit multi-modal distributions for the border locations. They thus yield far more detailed and useful information than the regression models that just predict 4 real values that correspond to estimations of the bounding box coordinates.
Furthermore, as a result of this, we  argue  that the task of learning to predict these probabilities is an easier one to accomplish.

To  implement the proposed localization model, we rely on  a convolutional neural network model, which we call \emph{LocNet}, whose architecture  is properly adapted such that the amount of parameters needed on the top fully connected layers is significantly reduced, thus making our LocNet model scalable with respect to the number of object categories.

Importantly, such a localization module can be easily incorporated into  many of the current state-of-the-art object detection systems~\cite{gidaris2015object,girshick2015fast,shaoqing2015faster},  helping them to significantly improve their localization performance. Here we use it in an iterative manner as part of a detection pipeline that  utilizes a recognition model for scoring candidate bounding boxes provided by the aforementioned localization module, and show that such an approach significantly boosts AP performance across a broad range of IoU thresholds. 

\textbf{Related work.} Most of the recent literature on object detection, 
treats the object localization problem at pre-recognition level by incorporating category-agnostic object proposal algorithms~\cite{van2011segmentation,zitnick2014edge,pinheiro2015learning,alexe2012measuring,krahenbuhl2014geodesic,krahenbuhllearning,APBMM2014,szegedy2013deep,szegedy2014scalable} that given an image, try to generate candidate boxes with high recall of the ground truth objects that they cover. 
Those proposals are later classified from a category-specific recognition model in order to create the final list of detections~\cite{girshick2014rich}.
Instead, in our work we focus on boosting the localization accuracy at post-recognition time, 
at which the improvements can be complementary to those obtained by improving the pre-recognition localization.
Till now, the work on this level has been limited to the bounding box regression paradigm that was first introduced from Felzenszwalb \etal~\cite{felzenszwalb2010object} and ever-since it has been used with success on most of the recent detection systems~\cite{girshick2014rich,girshick2015fast,shaoqing2015faster,sermanet2013overfeat,he2015spatial,yuting2015improving,zhu2015segdeepm,ren2015object,ouyang2014deepid}.
A regression model, 
given an initial candidate box that is loosely localized around an object, it tries to predict the coordinates of its ground truth bounding box.
Lately this model is enhanced by high capacity convolutional neural networks to further improve its localization capability~\cite{gidaris2015object,girshick2015fast,sermanet2013overfeat,shaoqing2015faster}.

To summarize, our  contributions are as follows:
\begin{itemize}
\item 
We cast the problem of localizing an object's bounding box as that of 
assigning probabilities on each row and column of a search region.
Those probabilities represent either the likelihood of each element (row or column) to belong on the inside of the bounding box or the likelihood to be one of the four borders of the object. 
Both of those cases is studied and compared with the bounding box regression model.
\item
To implement the above  model, we propose a properly adapted convolutional neural network architecture  that has  a reduced number of parameters and  results in an efficient and accurate object localization network (LocNet).
\item 
We extensively evaluate our approach on VOC2007~\cite{everingham2008pascal} and we show that it achieves a very significant improvement over the bounding box regression with respect to the mAP for IoU threshold of 0.7 and the COCO style of measuring the mAP.
It also offers an improvement with respect to the traditional way of measuring the mAP (\ie, for IoU $\geqslant 0.5$), achieving in this case $78.4\%$ 
and $74.78\%$ mAP on VOC2007~\cite{everingham2008pascal} and VOC2012~\cite{everingham2012pascal} 
test sets, which are the state-of-the-art at the time of writing this paper. 
Given those results we believe that our localization approach could  very well replace the existing bounding box regression paradigm in future object detection systems.  
\item
Finally we demonstrate that the detection accuracy of our system remains high even when it is given as input a set of \emph{sliding windows}, which proves that it is independent of bounding box proposal methods if the extra computational cost is ignored. 
\end{itemize}

The remainder of the paper is structured as follows:\ We describe our object detection methodology in \S\ref{sec:Methodology} and 
 then present our localization model in \S\ref{sec:Localization_model}. 
Implementation details and   experimental results are provided in \S\ref{sec:Implementation_Details} and \S\ref{sec:experimental_results} respectively. Finally, we conclude in \S\ref{sec:conclusions}.

\section{Object Detection Methodology}\label{sec:Methodology} 
\vspace{-15pt}
\begin{algorithm}\label{algo:detection}
\SetKwInOut{Input}{Input}
\SetKwInOut{Output}{Output}
\Input{Image $\textbf{I}$, initial set of candidate boxes $\textbf{B}^1$}
\Output{Final list of detections $\textbf{Y}$}

\For{$t \gets 1$ \textbf{to} $T$} {
$\textbf{S}^{t} \gets \textit{Recognition}(\textbf{B}^{t}| \textbf{I})$ \\

  \If{$t < T$} {
    $\textbf{B}^{t+1} \gets \textit{Localization}(\textbf{B}^{t}| \textbf{I})$ \\
  }
}
$\textbf{D} \gets \cup_{t=1}^{T}{\{\textbf{S}^{t},\textbf{B}^{t}\}}$ \\
$\textbf{Y} \gets \textit{PostProcess}(\textbf{D})$\\  
\caption{Object detection pipeline}\label{algo:detection}
\end{algorithm}

Our detection pipeline includes two basic components, the recognition and the localization models, integrated into an iterative scheme (see algorithm~\ref{algo:detection}). 
This scheme starts from an initial set of candidate boxes $\textbf{B}^1$ (which could be, \eg, either dense sliding windows~\cite{sermanet2013overfeat,papandreou2015modeling,redmon2015yolo,lenc2015r} or category-agnostic bounding box proposals~\cite{zitnick2014edge,van2011segmentation,shaoqing2015faster}) and on each iteration $t$ it uses the two basic components in the following way:
\begin{description}
\item[Recognition model:]
Given the current set of candidate boxes $\textbf{B}^{t} = \{B_{i}^{t}\}_{i=1}^{N_{t}}$, it assigns a confidence score to each of them $\{s_{i}^{t}\}_{i=1}^{N_{t}}$ that represents how likely it is for those boxes to be localized on an object of interest.
\item[Localization model:]
Given the current set of candidate boxes $\textbf{B}^{t} = \{B_{i}^{t}\}_{i=1}^{N_{t}}$, it generates a new set of candidate boxes $\textbf{B}^{t+1} = \{B_{i}^{t+1}\}_{i=1}^{N_{t+1}}$ such that those boxes they will be ``closer'' (\ie, better localized) on the objects of interest (so that they are probably scored higher from the recognition model).
\end{description}

In the end, the candidate boxes that were generated on each iteration from the localization model along with the confidences scores that were assigned to them from the recognition model are merged together and a post-processing step of non-max-suppression~\cite{felzenszwalb2010object} 
followed from bounding box voting~\cite{gidaris2015object} is applied to them.
The output of this post-processing step consists the detections set produced from our pipeline.
Both the recognition and the localization models are implemented as convolutional neural networks~\cite{lecun1989backpropagation} that lately have been empirically proven quite successful on computers vision tasks and especially those related to object recognition problems~\cite{simonyan2014very,krizhevsky2012imagenet,he2015delving,ioffe2015batch,szegedy2015going}. 
More details about our detection pipeline are provided in appendix~\ref{sec:pipeline_details}. 

Iterative object localization has also been explored before~\cite{caicedo2015active, gidaris2015object, gonzalez2015active, yoo2015attentionnet}. 
Notably, Gidaris and Komodakis~\cite{gidaris2015object} combine CNN-based regression with iterative localization while Caicedo et al.~\cite{caicedo2015active} and Yoo et al.~\cite{yoo2015attentionnet} attempt to localize an object by sequentially choosing one among a few possible actions that either transform the bounding box or stop the searching procedure. 

\section{Localization model}\label{sec:Localization_model}
\begin{figure}[t!]
\centering
\renewcommand{\figurename}{Figure}
\renewcommand{\captionlabelfont}{\bf}
\renewcommand{\captionfont}{\small} 
\centering
\includegraphics[width=0.5\textwidth]{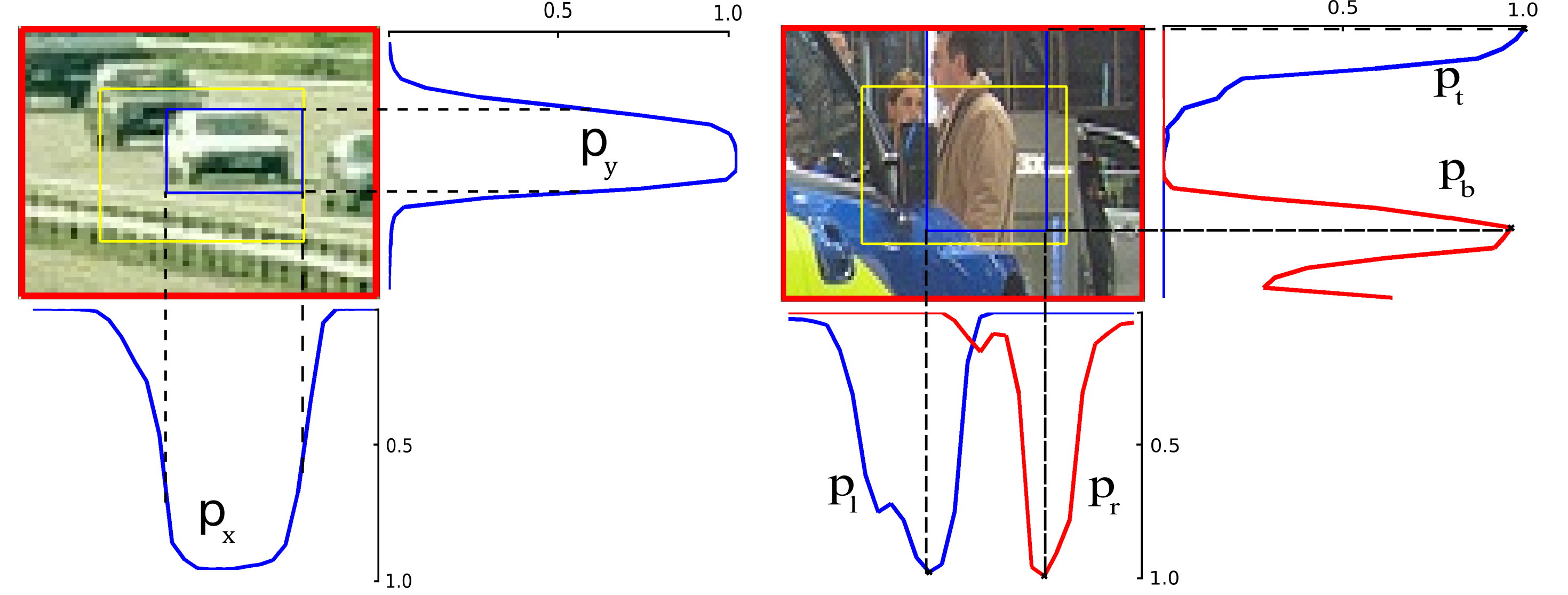}
\vspace{-1pt}
\caption{The posterior probabilities that our localization model yields given a region $R$. ~\emph{\textbf{Left Image:}} the in-out conditional probabilities that are assigned on each row ($p_{y}$) and column ($p_{x}$) of $R$. They are drawn with the blues curves on the right and on the  bottom  side of the search region. ~\emph{\textbf{Right Image:}} the conditional probabilities $p_l$, $p_r$, $p_t$, and $p_b$ of each column or row to be the left ($l$), right ($r$), top ($t$) and bottom ($b$) border of an object's bounding box. They are drawn with blue and red curves on the bottom and on the right side of the search region.}
\label{fig:Probabilities}        
\end{figure}

In this paper we focus on improving the localization model of this pipeline.
The  abstract work-flow that we use for this model is that it gets as input a candidate box $B$ in the image, it enlarges it by a factor $\gamma$\footnote{We use $\gamma=1.8$ in  all of the experiments.} to create a search region $R$ and then it returns a new candidate box that ideally will tightly enclose an object of interest in this region (see right column of Figure~\ref{fig:LocModelInOut}).

The crucial question is, of course, what is the most effective approach for constructing a model that is able to generate a good box prediction. One choice could be, for instance, to learn a regression function that directly predicts the 4 bounding box coordinates. However, we argue that this is not the most effective solution. Instead, we opt  for a different approach, which is detailed in the next section.
 
\subsection{Model predictions} \label{sec:model_predictions}

Given a search region $R$ and  object category $c$,
our object localization model  considers a division of $R$ in $M$   equal horizontal regions  (rows) as well as a  division of $R$ in $M$ equal vertical regions  (columns), and  outputs for each of them one or more conditional probabilities.
Each of these conditional probabilities is essentially a vector of the form $p^{R,c} = \{p(i|R,c)\}_{i=1}^{M}$ 
(hereafter we drop the $R$ and $c$ conditioned variables so as to reduce notational clutter). 
Two types of conditional probabilities are considered here:

\emph{\textbf{In-Out probabilities:}}
These are vectors ${p}_{x} \!=\! \{p_x(i)\}_{i=1}^{M}$ and ${p}_{y} = \{p_y(i)\}_{i=1}^{M}$ that  represent respectively the conditional probabilities of each column  and row of $R$  to be inside the bounding box of an object of category $c$ (see left part of Figure~\ref{fig:Probabilities}).
A row or column is considered to be inside a bounding box if at least part of the region corresponding to this row or column is inside this box.
For example, if $B^{gt}$ 
is  a ground truth bounding box with top-left coordinates  $(B^{gt}_l, B^{gt}_t)$ and bottom-right coordinates  $(B^{gt}_r, B^{gt}_b)$,\footnote{We actually assume that the ground truth bounding box is projected on the output domain of our model where the coordinates take integer values in the range $\{1,\ldots,M\}$. This is a necessary step for the definition of the target probabilities} then the In-Out probabilities   $p=\{p_x,p_y\}$ from the localization model should ideally  equal  to the following target probabilities $T = \{T_x, T_y\}$:
\[ \label{eq:fg_pred_eq_x}
  \forall i\in \{1,\ldots,M\},\ \ T_x(i)=\left\{
  \begin{array}{@{}ll@{}}
    1, & \text{if}\ B^{gt}_l \leq i \leq B^{gt}_r \\
    0, & \text{otherwise}
  \end{array}\right.\text{,}
\]
\[ \label{eq:fg_pred_eq_y}
  \forall i\in \{1,\ldots,M\},\ \ T_y(i)=\left\{
  \begin{array}{@{}ll@{}}
    1, & \text{if}\ B^{gt}_t \leq i \leq B^{gt}_b \\
    0, & \text{otherwise}
  \end{array}\right.\text{.}
\] 

\emph{\textbf{Border probabilities:}}
These are vectors
$p_{l}\! =\! \{p_l(i)\}_{i=1}^{M}$, $p_{r} = \{p_r(i)\}_{i=1}^{M}$, $p_{t} = \{p_t(i)\}_{i=1}^{M}$ and $p_{b} = \{p_b(i)\}_{i=1}^{M}$ that represent respectively the conditional probability of each column or row to be the left ($l$), right ($r$), top ($t$) and bottom ($b$) border  of the bounding box of an object of category $c$ (see right part of Figure~\ref{fig:Probabilities}).
In this case, the target probabilities $T = \{T_l, T_r, T_t, T_b\}$ that should   ideally be  predicted by the localization model for a ground truth bounding box $B^{gt}=(B^{gt}_l, B^{gt}_t, B^{gt}_r, B^{gt}_b)$ are given by
\[ \label{eq:side_pred_eq_log}
  \forall i\in \{1,\ldots,M\},\ \ T_{s}(i)=\left\{
  \begin{array}{@{}ll@{}}
     1, & \text{if}\ i = B^{gt}_s \\
     0, & \text{otherwise}
  \end{array}\right.\text{,}
\]
where $s \in \{l,r,t,b\}$. 
Note that we assume that the left and right border probabilities are independent and similarly for the top and bottom cases.

\subsubsection{Bounding box inference} Given the above output conditional probabilities, we  model the inference of the bounding box location 
$B=(B_l, B_t, B_r, B_b)$ 
using one of the following probabilistic models:

\emph{\textbf{In-Out ML:}} Maximizes the likelihood of the \emph{in-out} elements of $B$
\begin{align} \label{eq:posterior_prob}
L_{\textrm{in-out}}(B) = \prod_{i \in \{B_l,\ldots,B_r\}}{p_x(i)} \prod_{i \in \{B_t,\ldots,B_b\}}{p_y(i)}\nonumber\\ \prod_{i \notin  \{B_l,\ldots,B_r\}}{\tilde{p}_x(i)} \prod_{i \notin  \{B_t,\ldots,B_b\}}{\tilde{p}_y(i)} \text{,}
\end{align} 
where $\tilde{p}_x(i) = 1 - p_x(i)$ and $\tilde{p}_y(i) = 1 - p_y(i)$. The first two terms in the right hand of the equation represent the likelihood of the rows and columns of box $B$ (\emph{in}-elements) to be inside a ground truth bounding box and the last two terms the likelihood of the rows and columns that are not part of $B$ (\emph{out}-elements) to be outside  a ground truth bounding box.

\emph{\textbf{Borders ML:}} Maximizes the likelihood of the borders of box $B$:
\begin{equation} \label{eq:posterior_prob}{
L_{\textrm{borders}}(B) = p_l(B_l) \cdot p_t(B_t) \cdot p_r(B_r) \cdot p_b(B_b) \text{.}}
\end{equation} 

\emph{\textbf{Combined ML:}} It uses both types of probability distributions by maximizing the likelihood for both the \emph{borders} and the \emph{in-out} elements of $B$:
\begin{equation} \label{eq:posterior_prob}{
L_{\textrm{combined}}(B)=L_{\textrm{borders}}(B) \cdot L_{\textrm{in-out}}(B) \text{.}}
\end{equation} 

\subsubsection{Discussion}

\begin{figure}[h]
\centering
\renewcommand{\figurename}{Figure}
\renewcommand{\captionlabelfont}{\bf}
\renewcommand{\captionfont}{\small} 
\centering
\includegraphics[width=0.45\textwidth]{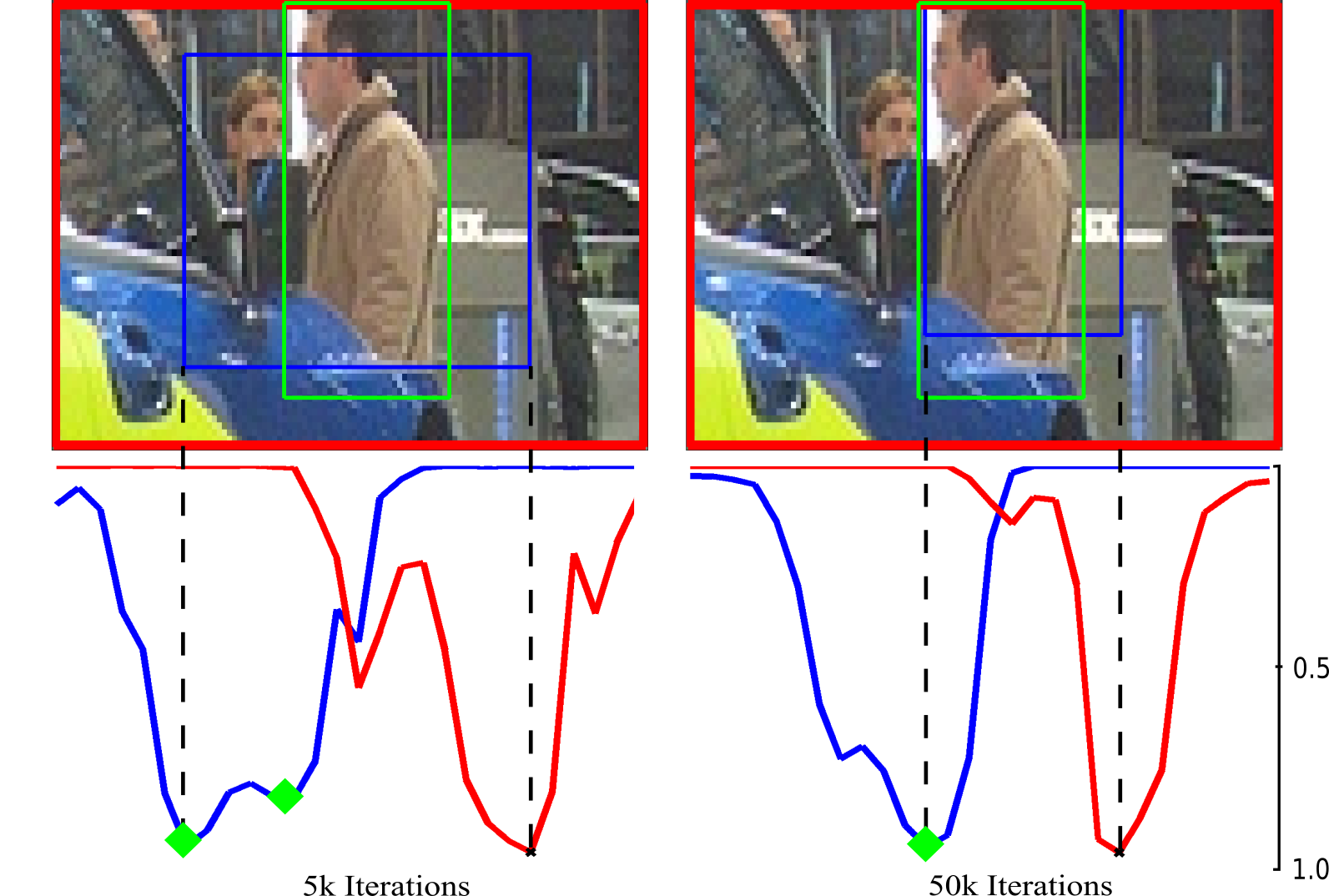}
\vspace{5pt}
\caption{
We show the evolution during training.
In the left image the green squares indicate the two highest modes of the left border probabilities predicted by a network trained only for a few iterations (5k).
Despite the fact that the highest one is erroneous, the network also maintains information for the correct mode. As training progresses (50k), this helps the network to correct its mistake and recover a correct left border(right image).}
\label{fig:Training}        
\end{figure}

\begin{figure}[h!]
\center
\renewcommand{\figurename}{Figure}
\renewcommand{\captionlabelfont}{\bf}
\renewcommand{\captionfont}{\small} 
        \begin{center}
        \includegraphics[width=0.45\textwidth]{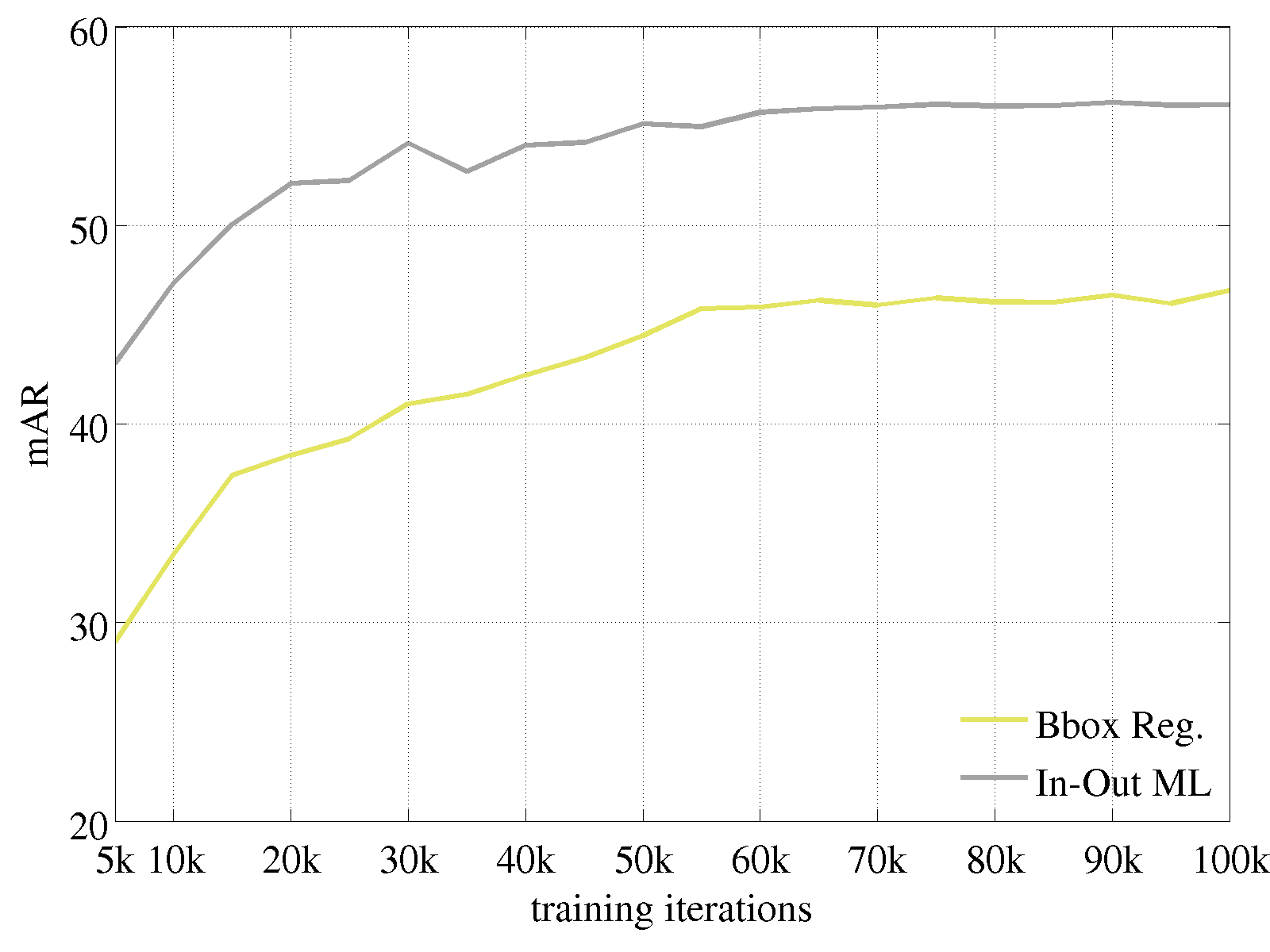}   
        \end{center}
\vspace{5pt}
\caption{\small{mAR as a function of the training iteration for the bounding box regression model (\emph{Bbox reg.}) and the \emph{In-Out ML} localization model. 
In order to create this plot, we created a small validation set of candidate boxes with a ground truth bounding box assigned on each of them, 
and during training given those candidates as input to the models we measure the mAR of the predicted boxes. We observe that the \emph{In-Out ML} localization model converges faster and to a higher mAR than the \emph{bounding box regression} localization model.}}
\label{tab:mAR_per_Iteration}
\end{figure}

The reason  we consider that the proposed   formulation of the problem of localizing an object's bounding box is superior  is because the In-Out or Border probabilities provide much more detailed and useful information regarding the location of a bounding box compared to the typical bounding box regression paradigm~\cite{felzenszwalb2010object}. In particular, in the later case the model simply directly predicts real values that corresponds to estimated bounding box coordinates but it does not provide, \eg, any confidence measure for these predictions. 
On the  contrary, our model provides a conditional probability for placing the four borders or the inside of an object's bounding box on each column and row of a search region $R$. As a result, it is perfectly capable of  handling also instances that  exhibit multi-modal conditional distributions (both during training and testing). During training, we argue that this makes the per row and per column probabilities much easier to be learned from a convolutional neural network that implements the model, than the bounding box regression task (\eg, see Figure~\ref{fig:Training}), thus helping the model to converge to a better training solution. 
Indeed, as we demonstrate, \eg, in Figure~\ref{tab:mAR_per_Iteration}, our CNN-based \emph{In-Out ML} localization model converges faster and on higher localization accuracy (measured with the mAR~\cite{hosang2015makes} metric) than a CNN-based bounding box regression model~\cite{girshick2015fast,gidaris2015object}. This behaviour was consistently observed in all of our proposed localization models.


\begin{figure}[t]
\centering
\renewcommand{\figurename}{Figure}
\renewcommand{\captionlabelfont}{\bf}
\renewcommand{\captionfont}{\small} 
\centering
\includegraphics[width=0.40\textwidth]{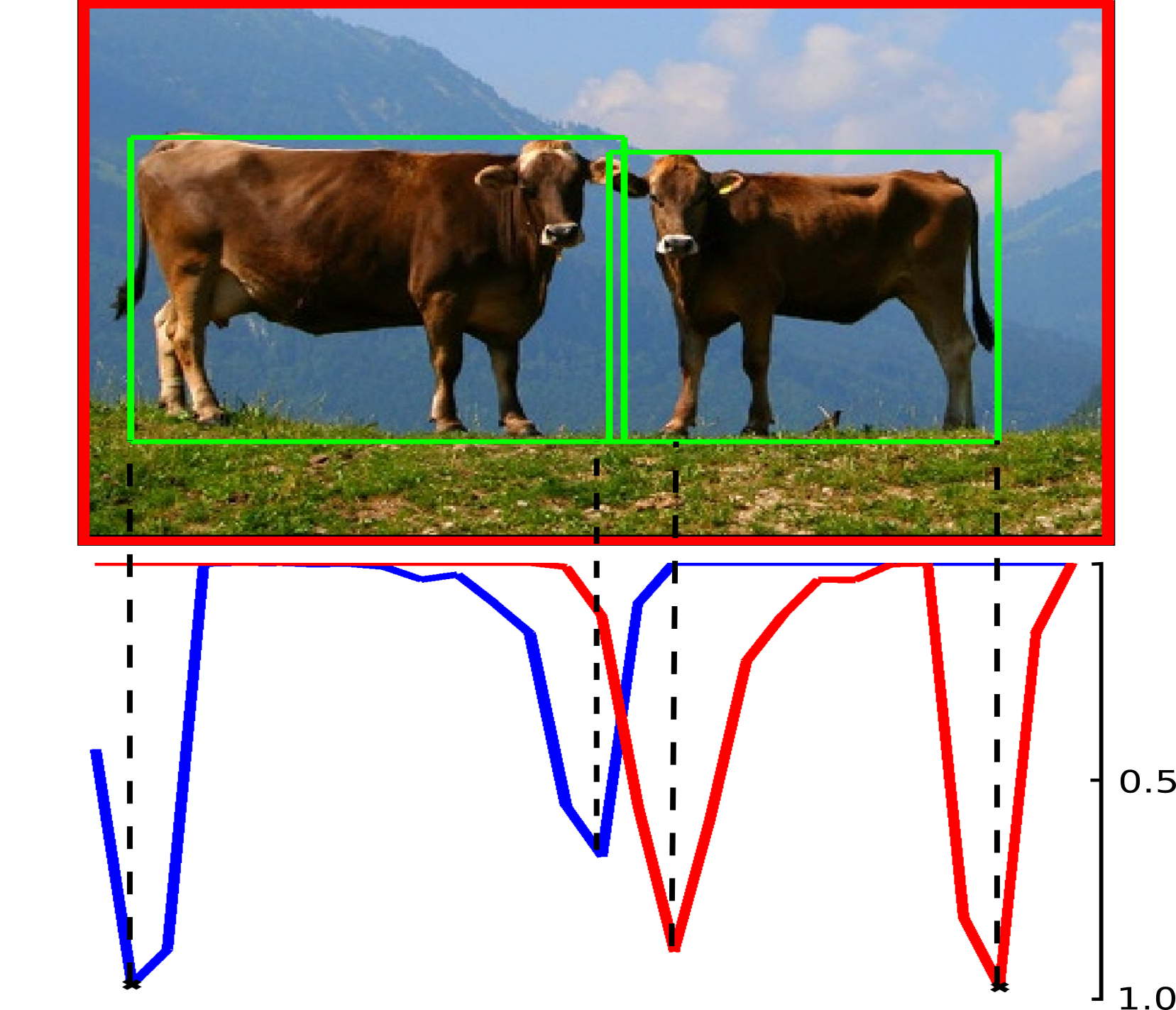}
\vspace{3pt}
\caption{
We depict the probabilities for the left (blue) and right (red) borders that a trained model yields for a region with two instances of the same class (cow).
The probability modes in this case can clearly indicate the presence of two instances.}
\label{fig:MultInstance}        
\end{figure}

Furthermore, during testing, these conditional distributions as we saw can be exploited in order to form probabilistic models for the inference of the bounding box coordinates.
In addition, they can  indicate the presence of a second instance inside the region $R$ and thus facilitate the localization of multiple adjacent instances, which is a difficult problem on object detection.
 In fact,  when visualizing, \eg, the border probabilities, we observed that this could have been  possible in several cases (\eg, see Figure~\ref{fig:MultInstance}). Although in this work we did not explore the possibility of utilizing a more advanced probabilistic model that  predicts $K>1$ boxes per region $R$, this can certainly be an interesting future addition to our method.  

Alternatively to our approach,
we could predict the probability of each pixel to belong on the foreground of an object,
as Pinheiro \etal~\cite{pinheiro2015learning} does. 
However, in order to learn such a type of model, pixel-wise instance segmentation masks are required during training, 
which in general is a rather tedious task to collect.
In contrary, for our model to learn those per row and per column probabilities, only bounding box annotations are required.
Even more, this independence is exploited in the design of the convolutional neural network that implements our model in order to keep the number of parameters of the prediction layers small (see \S~\ref{sec:loc_architecture}). 
This is significant for the scalability of our model with respect to the number of object categories since we favour category-specific object localization that has been shown to exhibit better localization accuracy~\cite{simonyan2014very}.

\subsection{LocNet network architecture} \label{sec:loc_architecture}
\begin{figure*}[t]
\center
\renewcommand{\figurename}{Figure}
\renewcommand{\captionlabelfont}{\bf}
\renewcommand{\captionfont}{\small} 
\centering
\includegraphics[width=\textwidth]{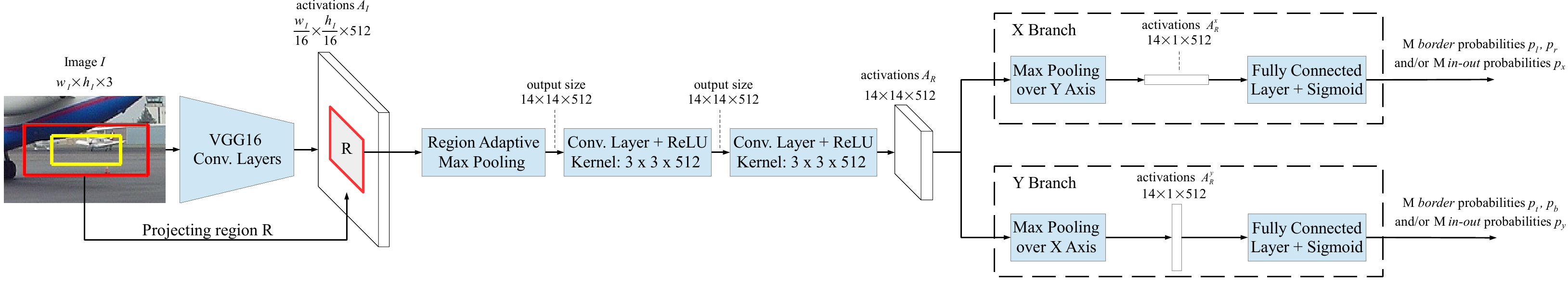}   
\vspace{0pt}
\caption{Visualization of the LocNet network architecture. In the input image, with yellow is drawn the candidate box $B$ and with red the search region $R$. In its output, the LocNet network yields probabilities for each of the $C$ object categories. The parameter $M$ that controls the output resolution is set to the value $28$ in our experiments. The convolutional layers of the VGG16-Net~\cite{simonyan2014very} that are being used in order to extract the image activations $A_I$ are those from conv1\_1 till conv5\_3. The new layers that are not derived from the VGG16-Net~\cite{simonyan2014very}, are randomly initialized with a Gaussian distribution with standard deviation of $0.001$ for the hidden layers and $0.01$ for the final fully connected layers.}
\label{fig:Architecture}        
\end{figure*}

Our localization model is implemented through the convolutional neural network that is visualized in Figure~\ref{fig:Architecture} and which is called LocNet.
The processing starts by forwarding the entire image $I$ (of size $w_I \times h_I$), through a sequence of convolutional layers (conv. layers of VGG16~\cite{simonyan2014very}) that outputs the $A_I$ activation maps (of size $\frac{w_I}{16} \times \frac{h_I}{16} \times 512$).  
Then, the region $R$ is projected on $A_I$ and the activations that lay inside it are cropped and pooled with a spatially adaptive max-pooling layer~\cite{he2015spatial}.
The resulting fixed size activation maps ($14 \times 14 \times 512$) are forwarded through two convolutional layers (of kernel size $3\times3\times512$), each followed by ReLU non-linearities, that yield the localization-aware activation maps $A_R$ of region $R$ (with dimensions size $14 \times 14 \times 512$).

At this point, given the activations $A_R$ the network yields the probabilities that were described in section~\S\ref{sec:model_predictions}. Specifically, the network is split into two branches, the \emph{X} and \emph{Y}, with each being dedicated for the predictions that correspond to the dimension ($x$ or $y$ respectively) that is assigned to it. 
Both start with a max-pool layer that aggregates the $A_{R}$ activation maps across the dimension perpendicular to the one dedicated to them, \ie,
\begin{equation} \label{eq:max_x}
A^{x}_{R}(i,f) = \max_{j} A_{R}(i,j,f), 
\end{equation}
\begin{equation} \label{eq:max_y}
A^{y}_{R}(j,f) = \max_{i} A_{R}(i,j,f) \text{,}
\end{equation} 
where $i$,$j$,and $f$ are the indices that span over the width, height, and feature channels of $A_R$ respectively.
The resulted activations $A^{x}_{R}$ and $A^{y}_{R}$ (both of size $14 \times 512$) efficiently encode the object location only across the dimension that their branch handles.
This aggregation process could also be described as marginalizing-out localization cues irrelevant for the dimension of interest.  
Finally, each of those aggregated features is fed into the final fully connected layer that is followed from sigmoid units in order to output the conditional probabilities of its assigned dimension.
Specifically, the \emph{X} branch outputs the $p_{x}$  and/or the $(p_{l}, p_{r})$ probability vectors whereas the \emph{Y} branch outputs the $p_{y}$  and/or the $(p_{t}, p_{b})$ probability vectors.
Despite the fact that
the last fully connected layers output category-specific predictions, 
their number of parameters remains relatively small due to the facts that: 1) they are applied on features of which the dimensionality has been previously drastically reduced due to the max-pooling layers of equations \ref{eq:max_x} and \ref{eq:max_y}, and 2) that each branch yields predictions only for a single dimension.

\subsection{Training} \label{sec:training}
During training, the network learns to map a search regions $R$ to the target probabilities $T$ that are conditioned on the object category $c$. Given a set of $N$ training samples $\{(R_k, T_k, c_k)\}_{k=1}^{N}$ the loss function that is minimised is
\begin{equation} \label{eq:total_loss_log}
L(\theta) = \frac{1}{N} \sum_{k=1}^{N}{l(\theta| R_k, T_k, c_k)}\textbf{,}
\end{equation}
where $\theta$ are the network parameters that are learned and $l(\theta| R, T, c)$ is the loss for one training sample. 

Both for the \emph{In-Out} and the \emph{Borders} probabilities we use the sum of binary logistic regression losses per row and column. 
Specifically, the per sample loss of the \emph{In-Out} case is:
\begin{equation} \label{eq:inout_loss_log}
\sum_{a \in\\ \{x,y\}}\sum_{i=1}^M T_a(i)\log(p_a(i))+\tilde{T}_a(i)\log(\tilde{p}_a(i))\,,
\end{equation}
and for the \emph{Borders} case is:
\begin{equation} \label{eq:inout_loss_log}
\mspace{-20mu}\sum_{s \in \{l,r,u,b\}}\sum_{i=1}^M\lambda^{+}T_s(i)\log(p_s(i))+\lambda^{-}\tilde{T}_s(i)\log(\tilde{p}_s(i))\,,
\end{equation}
where $\tilde{p} = 1-p$. In objective function \eqref{eq:inout_loss_log}, $\lambda^{+}$ and $\lambda^{-}$ represent the weightings of the losses   for misclassifying a border and a non-border element respectively. These are set as 
\[\lambda^{-} = 0.5 \cdot \frac{M}{M-1}\,,\ \ \lambda^{+} = (M-1)\cdot\lambda^{-}\,,\]
so as to balance the contribution on the loss of those two cases (note that $\tilde{T}_s(i)$ will be non-zero $M-1$ times more than ${T}_s(i)$). We observed that this leads to a model that yields more ``confident'' probabilities for the borders elements. 
For the \emph{Borders} case we also tried to use as loss function the Mean Square Error, while modifying the target probabilities to be Gaussian distributions around the border elements, but we did not observe an improvement in performance.

\section{Implementation details} \label{sec:Implementation_Details} 
\textbf{General:}
For the implementation code of our paper we make use of the Caffe framework~\cite{jia2014caffe}.
During training of all the models (both the localization and the recognition ones) we fine-tune only from the conv4\_1 convolutional layer and above. 
As training samples we use both selective search~\cite{van2011segmentation} and edge box~\cite{zitnick2014edge} proposals. 
Finally, both during training and testing we use a single image scale that is obtained after resizing the image such as its smallest dimension to be $600$ pixels.\\

\textbf{Proposed localization models (\emph{In-Out ML}, \emph{Borders ML}, \emph{Combined ML}):}
To create the training samples we take proposals of which the IoU with a ground truth bounding box is at least $0.4$, we enlarge them by a factor of $1.8$ in order to obtain the search regions $R$, and we assign to them the ground truth bounding box with which the original box proposal has the highest IoU in order to obtain the target bounding boxes and the corresponding target vectors $T$. This process is performed independently for each category. 
The parameter $M$ that controls the output resolution of our networks, is set to the value $28$.
For optimization we use stochastic gradient descend (SGD) with mini-batch size of 128 training candidate boxes. 
To speed up the training procedure we follow the paradigm of Fast-RCNN~\cite{girshick2015fast} and those 128 training candidate boxes are coming from only two images on each mini-batch.
The weight decay is set to $0.00005$ and the learning rate is set to $0.001$ and is reduced by a factor of $10$ after each $60k$ iterations. 
The overall training procedure is continued for up to $150k$ iterations and it takes around $1.5$ days in one NVIDIA Titan Black GPU.

\section{Experimental results} \label{sec:experimental_results}
\begin{figure*}[t!]
\center
\renewcommand{\figurename}{Figure}
\renewcommand{\captionlabelfont}{\bf}
\renewcommand{\captionfont}{\small} 

\begin{subfigure}[b]{0.45\textwidth}
\begin{center}
\includegraphics[width=\textwidth]{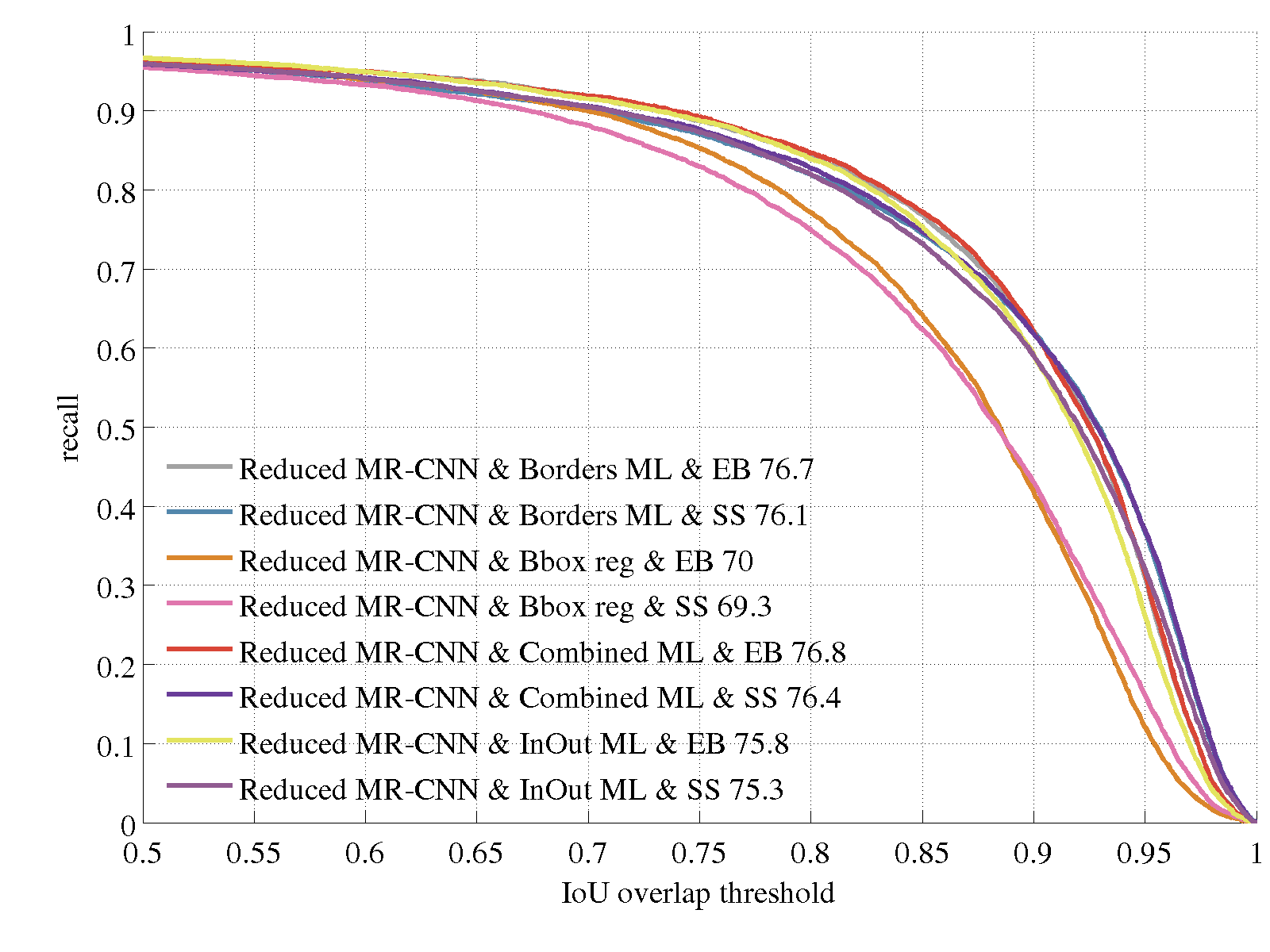}
\end{center}
\end{subfigure}
\begin{subfigure}[b]{0.45\textwidth}
\begin{center}
\includegraphics[width=\textwidth]{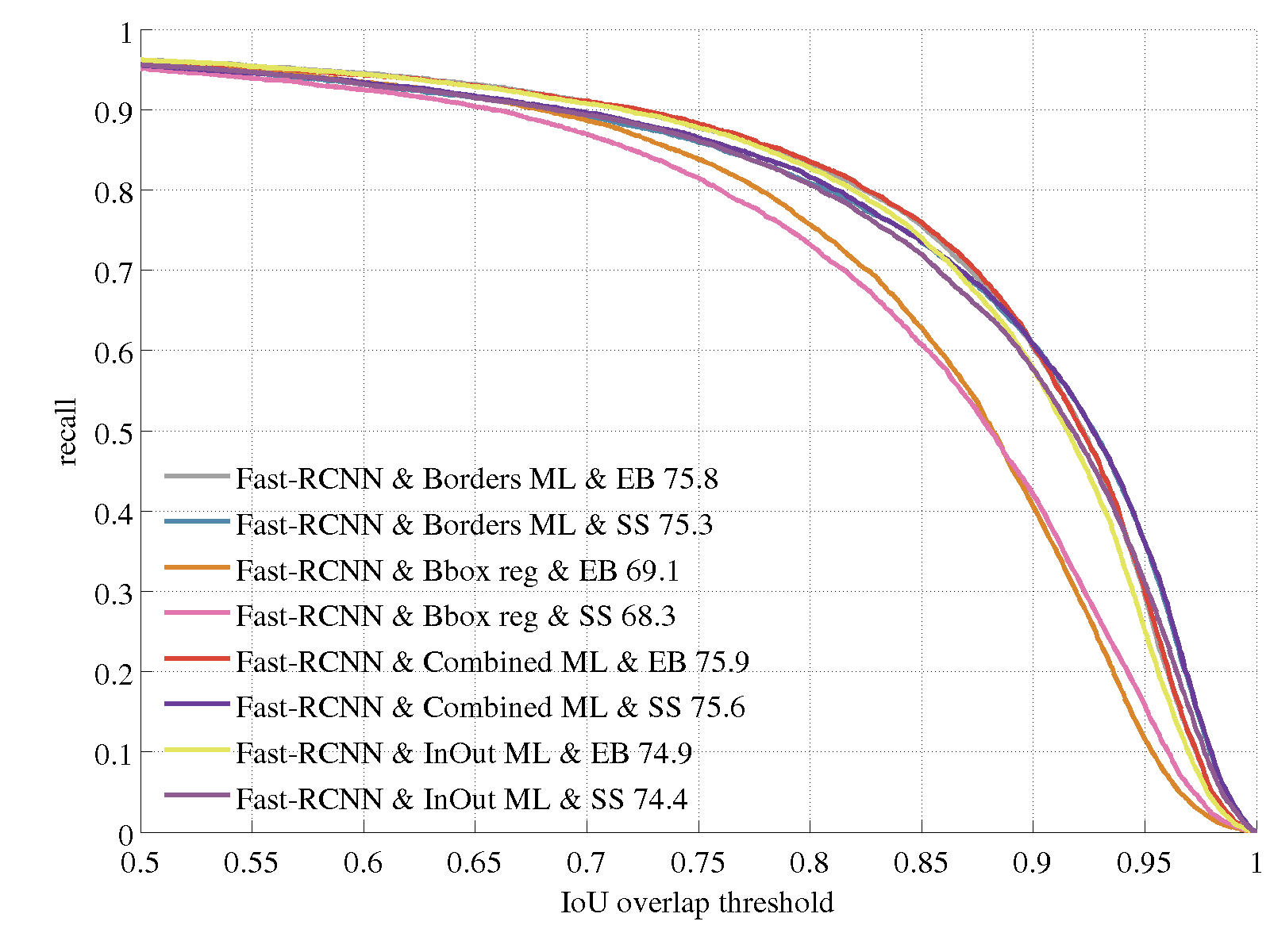}
\end{center}
\end{subfigure} \\
\begin{subfigure}[b]{0.45\textwidth}
\begin{center}
\includegraphics[width=\textwidth]{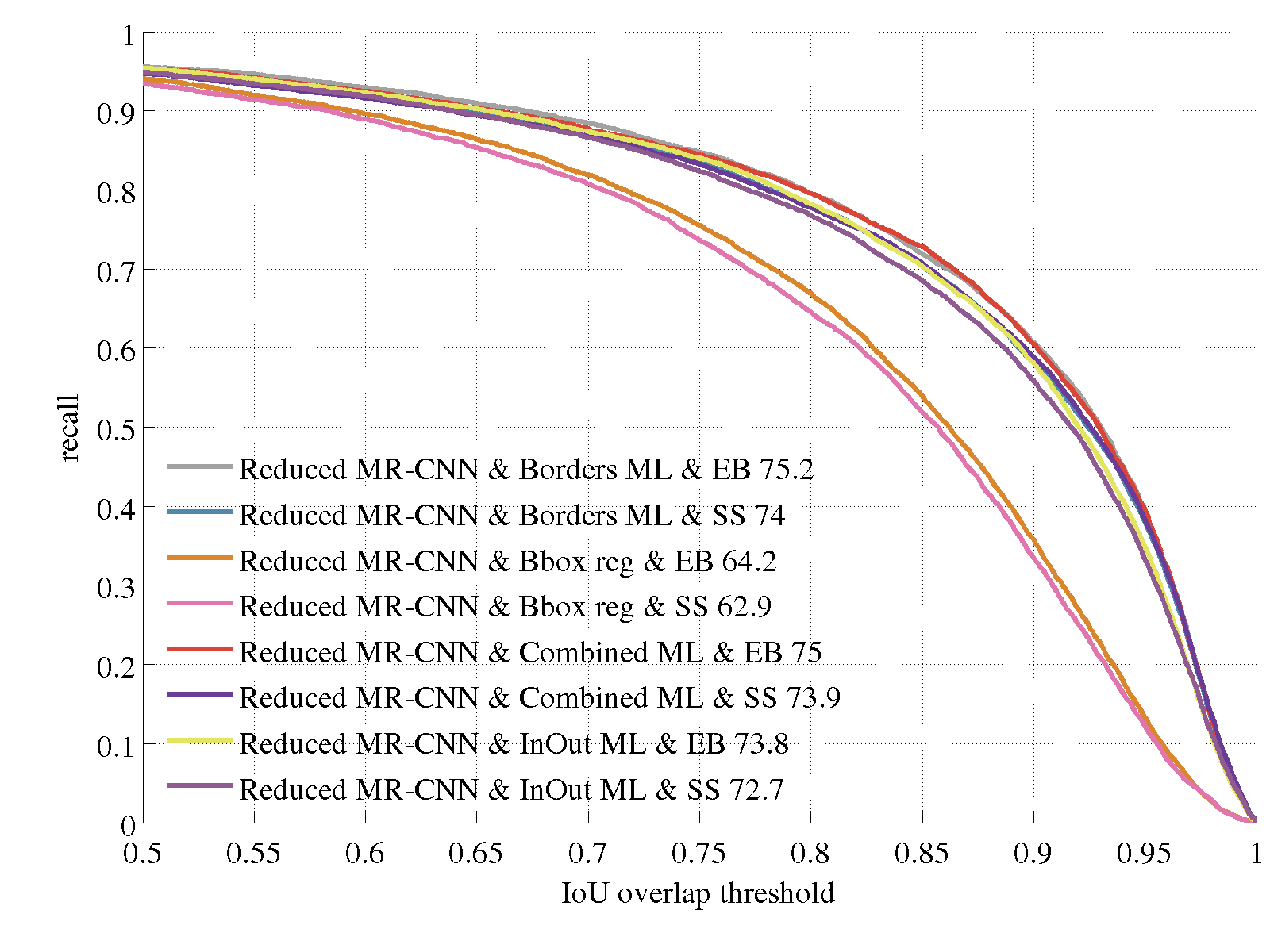}
\end{center}
\end{subfigure}
\begin{subfigure}[b]{0.45\textwidth}
\begin{center}
\includegraphics[width=\textwidth]{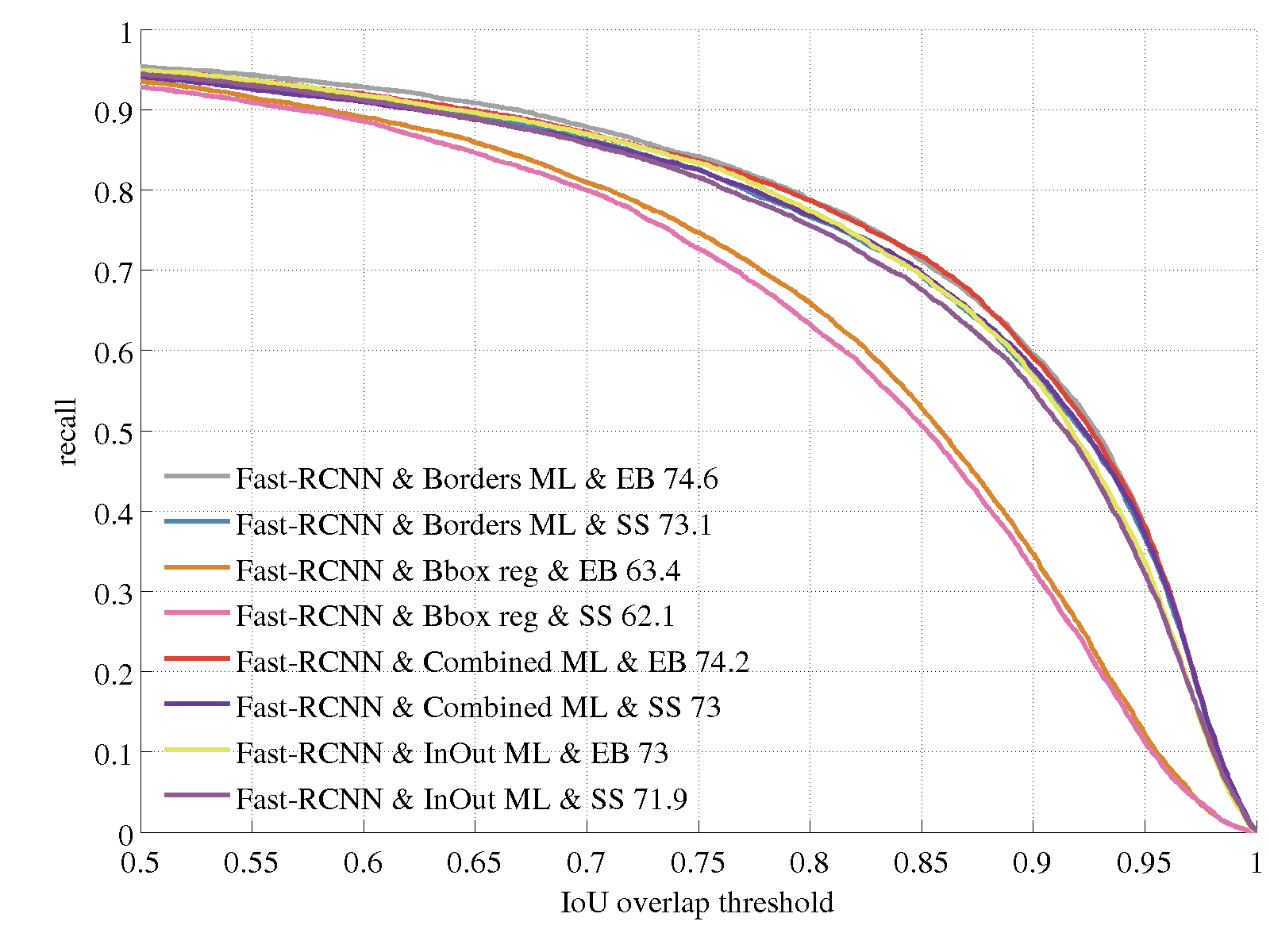}
\end{center}
\end{subfigure}      
\vspace{-1pt}
\caption{\small{Recall of ground truth bounding boxes as a function of the IoU threshold on PASCAL VOC2007 test set. Note that, because we perform class-specific localization the recall that those plots report is obtained after averaging the per class recalls. \emph{\textbf{Top-Left:}} Recalls for the \emph{Reduced MR-CNN} model after one iteration of the detection pipeline. \emph{\textbf{Bottom-Left:}} Recalls for the \emph{Reduced MR-CNN} model after four iterations of the detection pipeline.\emph{\textbf{Top-Right:}} Recalls for the \emph{Fast-RCNN} model after one iteration of the detection pipeline. \emph{\textbf{Bottom-Right:}} Recalls for the \emph{Fast-RCNN} model after four iterations of the detection pipeline.}}
\label{tab:Recall_IoU}
\vspace{10pt}
\end{figure*}
\begin{figure*}[t!]
\center
\renewcommand{\figurename}{Figure}
\renewcommand{\captionlabelfont}{\bf}
\renewcommand{\captionfont}{\small} 
\begin{subfigure}[b]{0.45\textwidth}
\begin{center}
\includegraphics[width=\textwidth]{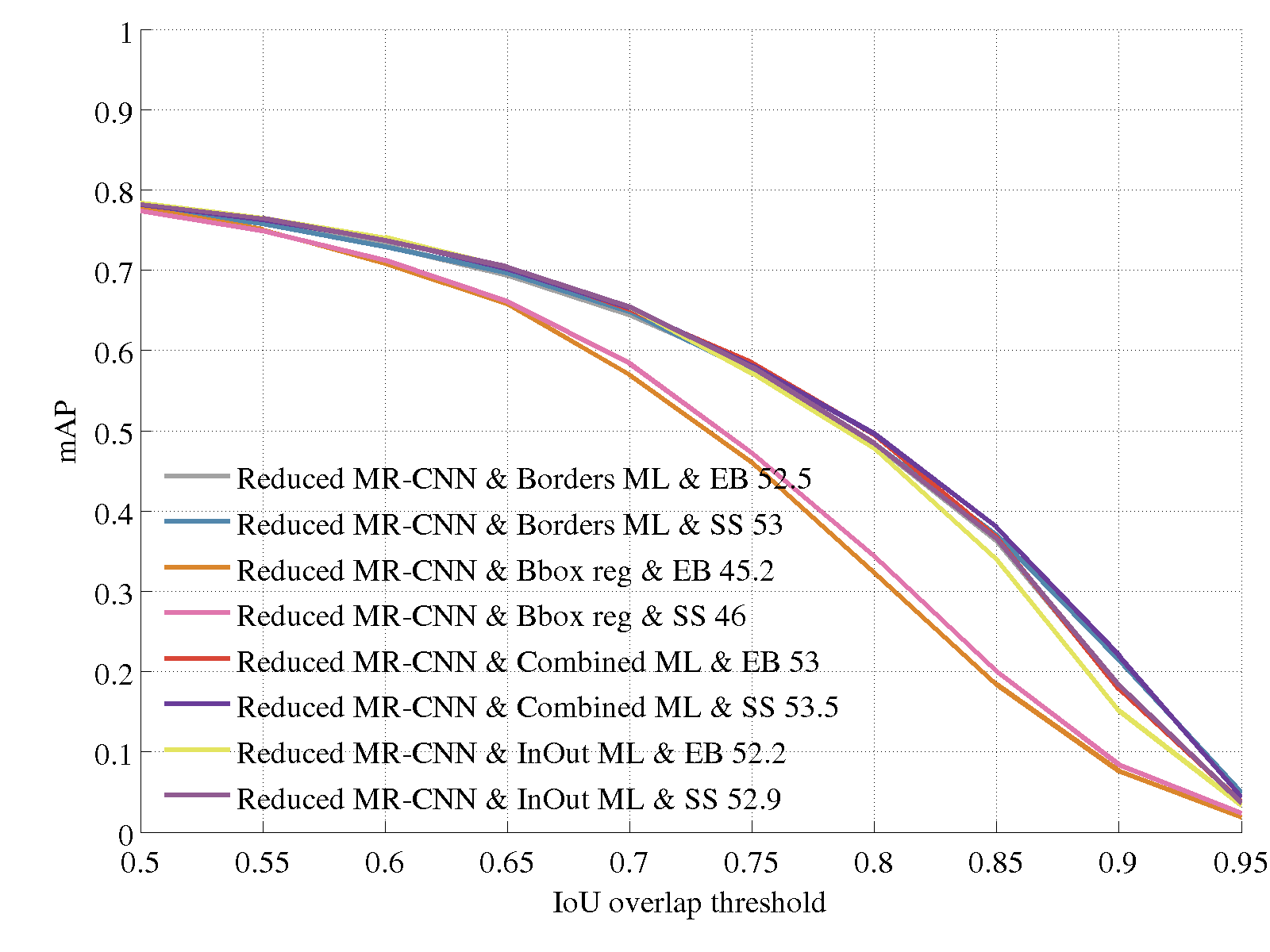}   
\end{center}
\end{subfigure}
\begin{subfigure}[b]{0.45\textwidth}
\begin{center}
\includegraphics[width=\textwidth]{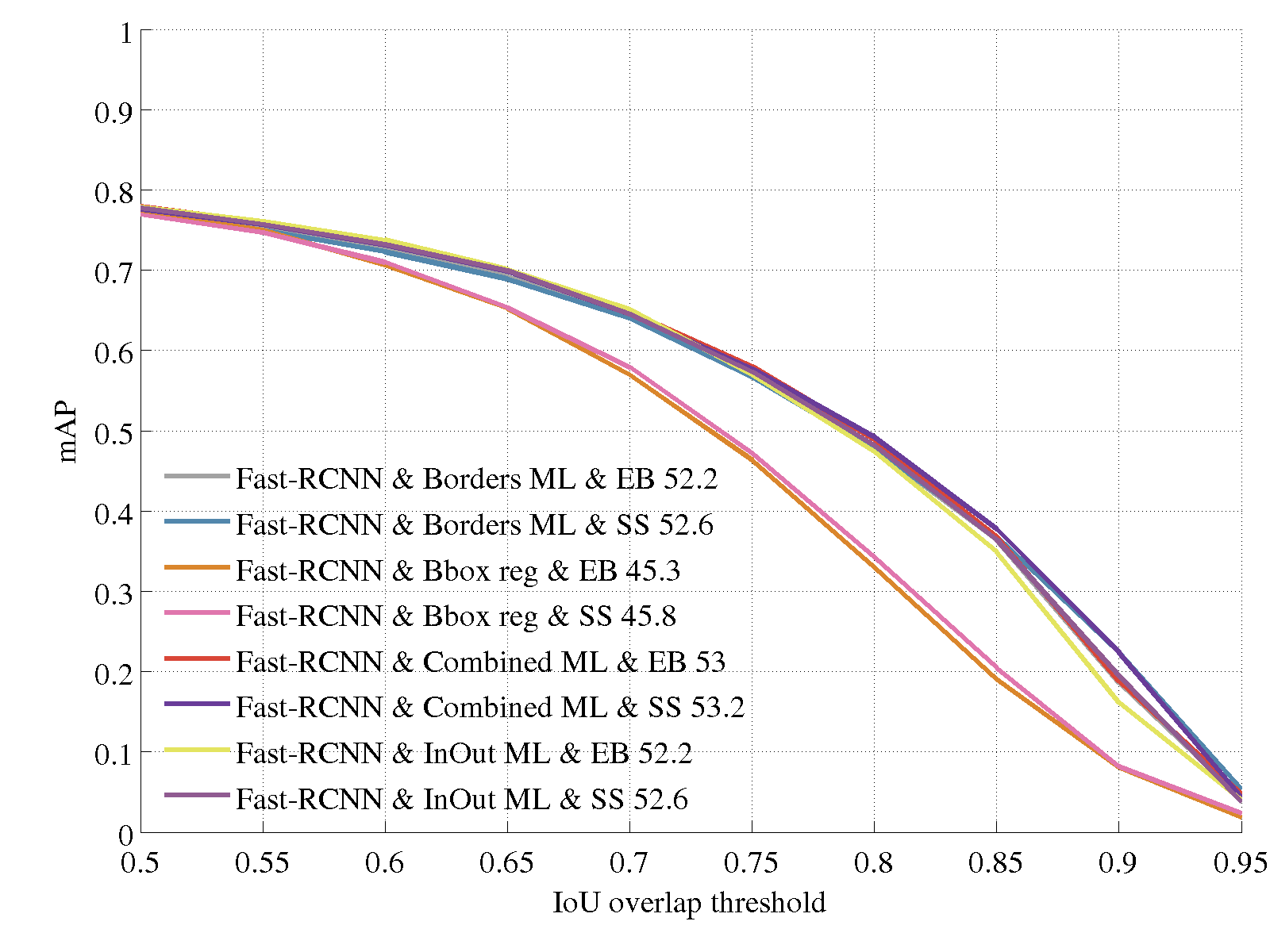}   
\end{center}
\end{subfigure} \\
\caption{\small{mAP as a function of the IoU threshold on PASCAL VOC2007 test set.
\emph{\textbf{Left plot:}} includes the configurations with the \emph{Reduced-MR-CNN} recognition model. \emph{\textbf{Right plot:}} includes the configurations with the \emph{Fast-RCNN} recognition model.}}
\label{tab:MRCNN_mAP_IoU}
\vspace{10pt}
\end{figure*}

We empirically evaluate our localization models on PASCAL VOC detection challenge~\cite{everingham2010pascal}. 
Specifically, we train all the recognition and localization models on VOC2007+2012 trainval sets and we test them on the VOC2007 test set. 
As baseline we use a CNN-based bounding box regression model~\cite{gidaris2015object} (see appendices~\ref{sec:bbox_reg}, ~\ref{sec:train_loc_models}, and ~\ref{sec:exps_size}). The remaining  components of the detection pipeline include\\
\textbf{\emph{Initial set of candidate boxes:}} 
We examine three alternatives for generating the initial set of candidate boxes: the Edge Box algorithm~\cite{zitnick2014edge} (\emph{EB}), the Selective Search algorithm (\emph{SS}), and a sliding windows scheme. In Table~\ref{tab:recall_proposals} we provide the recall statistics of the those bounding box proposal methods.\\
\textbf{\emph{Recognition model:}}
For the recognition part of the detection system we use either the \emph{Fast-RCNN~\cite{girshick2015fast}} or the \emph{MR-CNN~\cite{gidaris2015object}} recognition models.
During implementing the latter one, we  performed several simplifications on its architecture and thus we call the resulting model \emph{Reduced-MR-CNN} (those modifications are detailed in appendix~\ref{sec:rec_models}). 
The Fast-RCNN and Reduced-MR-CNN models are trained using both selective search and edge box proposals and as top layer they have class-specific linear SVMs~\cite{girshick2014rich}. 

First, we examine the performance of our approach with respect to localization (\S\ref{sec:loc_per}) and detection (\S\ref{sec:det_per}) accuracy. 
Then we report the detection accuracy of our approach for the sliding windows case (\S\ref{sec:sliding_window}). 
Finally, we provide preliminary results of our approach on COCO detection challenge in~\S\ref{sec:exps_coco} and qualitative results in~\S\ref{sec:qualitative}.

\begin{table}[t!]
\centering
\renewcommand{\figurename}{Table}
\renewcommand{\captionlabelfont}{\bf}
\renewcommand{\captionfont}{\small} 
\resizebox{0.45\textwidth}{!}{
{\setlength{\extrarowheight}{2pt}\scriptsize
{\begin{tabular}{l <{\hspace{-0.3em}}|>{\hspace{-0.5em}} l || >{\hspace{-0.5em}}c | >{\hspace{-0.5em}}c | >{\hspace{-0.5em}}c }
\hline
\multirow{2}{1.8cm}{Initial set of candidate boxes} &  \multirow{2}{*}{Number} & \multicolumn{3}{c}{Recall}\\
\multicolumn{1}{c|}{} & \multicolumn{1}{c||}{}& IoU$\geq$0.5 & IoU$\geq$0.7 & mAR\\
\hline
\emph{Sliding Windows}  & around $10$k & 0.920 & 0.389 & 0.350\\
\emph{Edge Box}         & around $2$k  & 0.928 & 0.755 & 0.517\\
\emph{Sel. Search}      & around $2$k  & 0.936 & 0.687 & 0.528\\
\hline
\end{tabular}}}}
\vspace{3pt}
\caption{Recall statistics on VOC2007 test set of the box proposals methods that we use in our work in order to generate the initial set of candidate boxes.}
\label{tab:recall_proposals}
\end{table}
\subsection{Localization performance} \label{sec:loc_per}

We first evaluate merely the localization performance of our models, thus  ignoring in this case the recognition aspect of the detection problem. 
For that purpose we report the recall that the examined models achieve.
Specifically, in Figure~\ref{tab:Recall_IoU} we provide the recall as a function of the IoU threshold for the candidate boxes generated on the first iteration and the last iteration of our detection pipeline. Also, in the legends of these figures we report the average recall (AR)~\cite{hosang2015makes} that each model achieves. 
Note that, given the set of initial candidate boxes and the recognition model, the input to the iterative localization mechanism is exactly the same and thus any difference on the recall is solely due to the localization capabilities of the models.
We observe that for IoU thresholds above $0.65$, the proposed models achieve higher recall than bounding box regression and that this improvement is actually increased with more iterations of the localization module. 
Also, the AR of our proposed models is on average 6 points higher than bounding box regression.

\subsection{Detection performance} \label{sec:det_per}

\begin{table*}
\centering
\renewcommand{\figurename}{Table}
\renewcommand{\captionlabelfont}{\bf}
\renewcommand{\captionfont}{\small} 
\resizebox{0.7\textwidth}{!}{
{\setlength{\extrarowheight}{2pt}\scriptsize
{\begin{tabular}{l <{\hspace{-0.3em}}|>{\hspace{-0.5em}} l | >{\hspace{-0.5em}} l || >{\hspace{-0.5em}}c | >{\hspace{-0.5em}}c | >{\hspace{-0.5em}}c }
\hline
\multicolumn{3}{c||}{Detection Pipeline} & \multicolumn{3}{c}{mAP}\\
\hline
Localization & Recognition & Initial Boxes & IoU $\geq$ 0.5 & IoU $\geq$ 0.7 & \emph{COCO} style\\
\hline
--                         & \emph{Reduced-MR-CNN} & \emph{$2$k Edge Box} & 0.747 & 0.434 & 0.362\\
\emph{InOut ML}    & \emph{Reduced-MR-CNN} & \emph{$2$k Edge Box} & 0.783 & \textbf{0.654} & 0.522\\
\emph{Borders ML}  & \emph{Reduced-MR-CNN} & \emph{$2$k Edge Box} & 0.780 & 0.644 & 0.525\\
\emph{Combined ML} & \emph{Reduced-MR-CNN} & \emph{$2$k Edge Box} & \textbf{0.784} & 0.650 & 0.530\\
\emph{Bbox reg.}   & \emph{Reduced-MR-CNN} & \emph{$2$k Edge Box} & 0.777 & 0.570 & 0.452\\
\hline
--                         & \emph{Reduced-MR-CNN} & \emph{$2$k Sel. Search} & 0.719 & 0.456 & 0.368\\
\emph{InOut ML}    & \emph{Reduced-MR-CNN} & \emph{$2$k Sel. Search} & 0.782 & \textbf{0.654} & 0.529\\
\emph{Borders ML}  & \emph{Reduced-MR-CNN} & \emph{$2$k Sel. Search} & 0.777 & 0.648 & 0.530\\
\emph{Combined ML} & \emph{Reduced-MR-CNN} & \emph{$2$k Sel. Search} & 0.781 & 0.653 & \textbf{0.535}\\
\emph{Bbox reg.}   & \emph{Reduced-MR-CNN} & \emph{$2$k Sel. Search} & 0.774 & 0.584 & 0.460\\
\hline
--                         & \emph{Fast-RCNN} & \emph{$2$k Edge Box} & 0.729 & 0.427 & 0.356\\
\emph{InOut ML}    & \emph{Fast-RCNN} & \emph{$2$k Edge Box} & 0.779 & 0.651 & 0.522\\
\emph{Borders ML}  & \emph{Fast-RCNN} & \emph{$2$k Edge Box} & 0.774 & 0.641 & 0.522\\
\emph{Combined ML} & \emph{Fast-RCNN} & \emph{$2$k Edge Box} & 0.780 & 0.648 & 0.530\\
\emph{Bbox reg.}   & \emph{Fast-RCNN} & \emph{$2$k Edge Box} & 0.773 & 0.570 & 0.453\\
\hline
--                         & \emph{Fast-RCNN} &  \emph{$2$k Sel. Search} & 0.710 & 0.446 & 0.362\\
\emph{InOut ML}    & \emph{Fast-RCNN} &  \emph{$2$k Sel. Search} & 0.777 & 0.645 & 0.526\\
\emph{Borders ML}  & \emph{Fast-RCNN} &  \emph{$2$k Sel. Search} & 0.772 & 0.640 & 0.526\\
\emph{Combined ML} & \emph{Fast-RCNN} &  \emph{$2$k Sel. Search} & 0.775 & 0.645 & 0.532\\
\emph{Bbox reg.}   & \emph{Fast-RCNN} &  \emph{$2$k Sel. Search} & 0.769 & 0.579 & 0.458\\
\hline
\end{tabular}}}}
\vspace{3pt}
\caption{\small{mAP results on VOC2007 test set for IoU thresholds of 0.5 and 0.7 as well as the COCO style mAP that averages the traditional AP for various IoU thresholds between 0.5 and 1 (specifically the thresholds 0.5:0.05:95 are being used). 
The hyphen symbol (--) indicates that the localization model was not used at all and that the pipeline ran only for $T=1$ iteration.
The rest entries are obtained after running the detection pipeline for $T=4$ iterations.}}
\label{tab:mAP_voc2007_test}
\vspace{-5pt}
\end{table*}

\begin{table*}[t!]
\centering
\renewcommand{\figurename}{Table}
\renewcommand{\captionlabelfont}{\bf}
\renewcommand{\captionfont}{\small} 
\resizebox{\textwidth}{!}{
{\setlength{\extrarowheight}{2pt}\scriptsize
{\begin{tabular}{l <{\hspace{-0.3em}}|>{\hspace{-0.5em}} l |>{\hspace{-0.5em}} l | >{\hspace{-0.5em}}c >{\hspace{-1em}}c >{\hspace{-1em}}c >{\hspace{-1em}}c >{\hspace{-1em}}c >{\hspace{-1em}}c >{\hspace{-1em}}c >{\hspace{-1em}}c >{\hspace{-1em}}c >{\hspace{-1em}}c >{\hspace{-1em}}c >{\hspace{-1em}}c >{\hspace{-1em}}c >{\hspace{-1em}}c >{\hspace{-1em}}c >{\hspace{-1em}}c >{\hspace{-1em}}c >{\hspace{-1em}}c >{\hspace{-1em}}c >{\hspace{-1em}}c <{\hspace{-0.3em}}| >{\hspace{-0.3em}}c}
\hline
Year & Metric & Approach &  areo & bike & bird & boat & bottle & bus & car & cat & chair & cow & table & dog & horse & mbike & person & plant & sheep & sofa & train & tv & mean \\
\hline
2007 & IoU $\geq$ 0.5 & \emph{Reduced-MR-CNN \& Combined ML \& EB} & 0.804 & 0.855 & 0.776 & 0.729 & 0.622 & 0.868 & 0.875 & 0.886 & 0.613 & 0.860 & 0.739 & 0.861 & 0.870 & 0.826 & 0.791 & 0.517 & 0.794 & 0.752 & 0.866 & 0.777 & 0.784 \\ 
2007 & IoU $\geq$ 0.7 & \emph{Reduced-MR-CNN \& In Out ML \& EB} & 0.707 & 0.742 & 0.622 & 0.481 & 0.452 & 0.840 & 0.747 & 0.786 & 0.429 & 0.730 & 0.670 & 0.754 & 0.779 & 0.669 & 0.581 & 0.309 & 0.655 & 0.693 & 0.736 & 0.690 & 0.654 \\ 
2007 & COCO style & \emph{Reduced-MR-CNN \& Combined ML \& SS} & 0.580 & 0.603 & 0.500 & 0.413 & 0.367 & 0.703 & 0.631 & 0.661 & 0.357 & 0.581 & 0.500 & 0.620 & 0.625 & 0.545 & 0.494 & 0.269 & 0.522 & 0.579 & 0.602 & 0.555 & 0.535 \\ 
\hline
2012 & IoU $\geq$ 0.5 & \emph{Reduced-MR-CNN \& In Out ML \& EB} & 0.863 & 0.830 & 0.761 & 0.608 & 0.546 & 0.799 & 0.790 & 0.906 & 0.543 & 0.816 & 0.620 & 0.890 & 0.857 & 0.855 & 0.828 & 0.497 & 0.766 & 0.675 & 0.832 & 0.674 & 0.748\\ 
2012 & IoU $\geq$ 0.5 & \emph{Reduced-MR-CNN \& Borders ML \& EB} & 0.865 & 0.827 & 0.755 & 0.602 & 0.535 & 0.791 & 0.785 & 0.902 & 0.533 & 0.800 & 0.607 & 0.886 & 0.857 & 0.848 & 0.826 & 0.496 & 0.765 & 0.673 & 0.831 & 0.676 & 0.743\\ 
2012 & IoU $\geq$ 0.5 & \emph{Reduced-MR-CNN \& Combined ML \& EB} & 0.866 & 0.834 & 0.765 & 0.604 & 0.544 & 0.798 & 0.786 & 0.902 & 0.546 & 0.810 & 0.618 & 0.889 & 0.857 & 0.847 & 0.828 & 0.498 & 0.763 & 0.678 & 0.830 & 0.679 & 0.747\\ 
\hline
\end{tabular}}}}
\vspace{1pt}
\caption{Per class AP results on VOC2007 and VO2012 test sets.}
\label{tab:final_system_voc2007_test_extra_data}
\end{table*}

Here we evaluate the detection performance of the examined localization models when plugged into the detection pipeline that was described in section~\S\ref{sec:Methodology}. 
In Table~\ref{tab:mAP_voc2007_test} we report the mAP on VOC2007 test set for IoU thresholds of $0.5$ and $0.7$ as well as the COCO style of mAP that averages the traditional mAP over various IoU thresholds between $0.5$ and $1.0$. The results that are reported are obtained after running the detection pipeline for $T=4$ iterations.
We observe that the proposed \emph{InOut ML}, \emph{Borders ML}, and \emph{Combined ML} localization models offer a significant boost on the mAP for IoU $\geq0.7$ and the COCO style mAP, relative to the bounding box regression model (\emph{Bbox reg.}) under all the tested cases. The improvement on both of them is on average $7$ points. Our models also improve for the mAP with IoU$\geq0.5$ case but with a smaller amount (around $0.7$ points). 
In Figure~\ref{tab:MRCNN_mAP_IoU} we plot the mAP as a function of the IoU threshold.
We can observe that the improvement on the detection performance thanks to the proposed localization models starts to clearly appear on the $0.65$ IoU threshold and then grows wider till the $0.9$.
In Table~\ref{tab:final_system_voc2007_test_extra_data} we provide the per class AP results on VOC2007 for the best approach on each metric.
In the same table we also report the AP results on VOC2012 test set but only for the IoU $\geq$ 0.5 case since this is the only metric that the evaluation server provides. In this dataset we achieve mAP of $74.8\%$ which is the state-of-the-art at the time of writing this paper (6/11/2015).
Finally, in Figure~\ref{fig:MRCNN_mAP_IoU_per_Iter} we examine the detection performance behaviour with respect to the number of iterations used by our pipeline. 
We observe that as we increase the number of iterations, the mAP for high IoU thresholds (e.g. IoU $\geq0.8$) continues to improve while for lower thresholds the improvements stop on the first two iterations. 

\begin{table*}[t!]
\centering
\renewcommand{\figurename}{Table}
\renewcommand{\captionlabelfont}{\bf}
\renewcommand{\captionfont}{\small} 
\resizebox{0.7\textwidth}{!}{
{\setlength{\extrarowheight}{2pt}\scriptsize
{\begin{tabular}{l <{\hspace{-0.3em}}|>{\hspace{-0.5em}} l | >{\hspace{-0.5em}} l || >{\hspace{-0.5em}}c | >{\hspace{-0.5em}}c | >{\hspace{-0.5em}}c }
\hline
\multicolumn{3}{c||}{Detection Pipeline} & \multicolumn{3}{c}{mAP}\\
\hline
Localization & Recognition & Initial Boxes & IoU $\geq$ 0.5 & IoU $\geq$ 0.7 & \emph{COCO} style\\
\hline
--                         & \emph{Reduced-MR-CNN} & \emph{$10$k Sliding Windows} & 0.617 & 0.174 & 0.227\\
\emph{InOut ML}    & \emph{Reduced-MR-CNN} & \emph{$10$k Sliding Windows} & 0.770 & 0.633 & 0.513\\
\emph{Borders ML}  & \emph{Reduced-MR-CNN} & \emph{$10$k Sliding Windows} & 0.764 & 0.626 & 0.513\\
\emph{Combined ML} & \emph{Reduced-MR-CNN} & \emph{$10$k Sliding Windows} & 0.773 & 0.639 & 0.521\\
\emph{Bbox reg.}   & \emph{Reduced-MR-CNN} & \emph{$10$k Sliding Windows} & 0.761 & 0.550 & 0.436\\
\hline
\end{tabular}}}}
\vspace{1pt}
\caption{\small{mAP results on VOC2007 test set when using \emph{$10$k sliding windows} as initial set of candidate boxes. 
In order to generate the sliding windows we use the publicly available code that accompanies the work of Hosang \etal~\cite{hosang2015makes} that includes a sliding window implementation inspired by \emph{BING}~\cite{cheng2014bing,zhao2014cracking}).}}
\label{tab:mAP_voc2007_test_sliding_windows}
\end{table*}
\begin{figure}[t!]
\center
\renewcommand{\figurename}{Figure}
\renewcommand{\captionlabelfont}{\bf}
\renewcommand{\captionfont}{\small} 
        \begin{center}
        \includegraphics[width=0.45\textwidth]{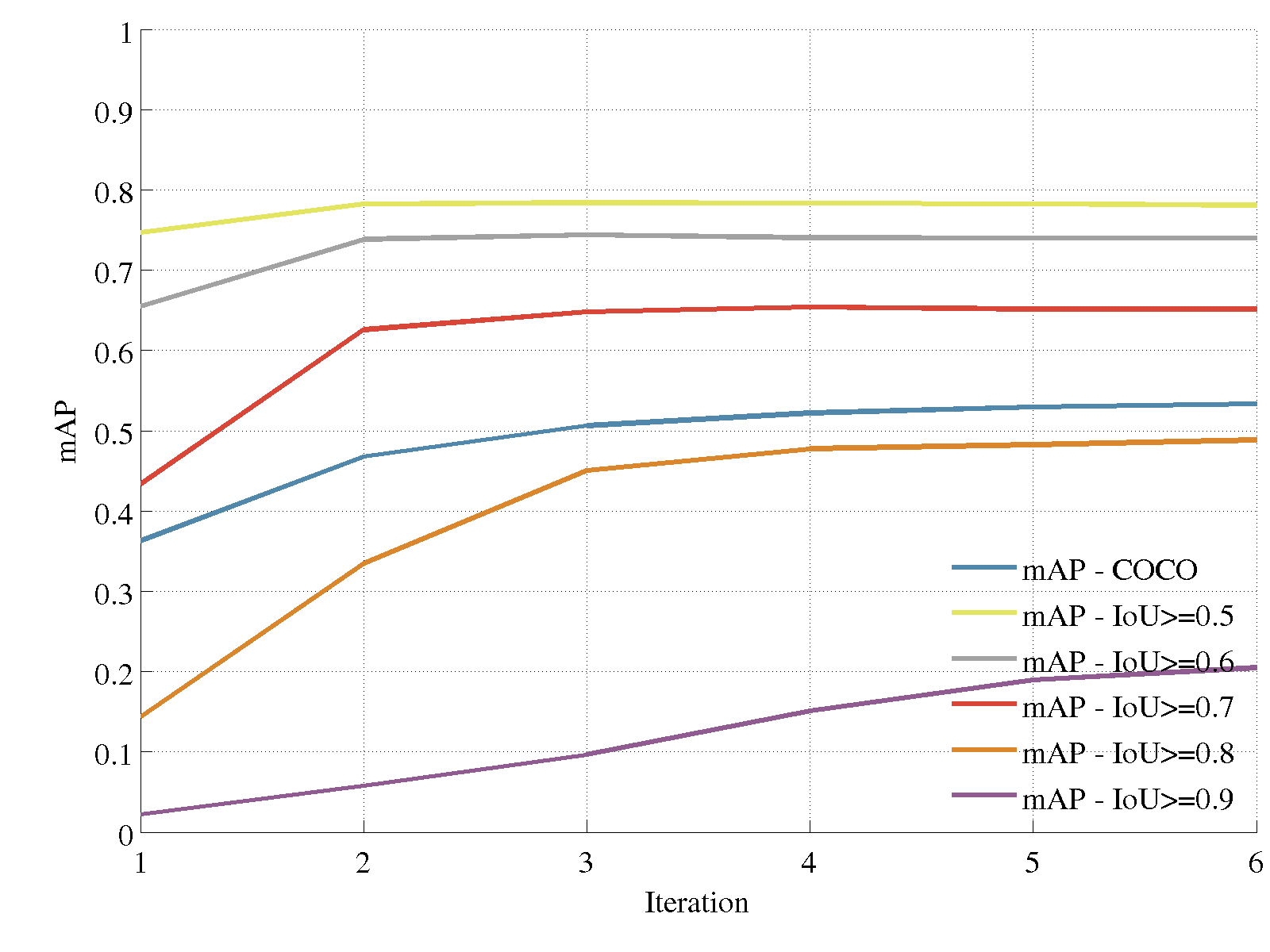}   
        \end{center}
\caption{\small{Plot of the mAP as a function of the iterations number of our detection pipeline on VOC2007 test set. To generate this plot we used the \emph{Reduced-MR-CNN} recognition model with the ~\emph{In-Out ML} localization model and Edge Box proposals.}} 
\label{fig:MRCNN_mAP_IoU_per_Iter}
\end{figure}

\subsection{Sliding windows as initial set of candidate boxes} \label{sec:sliding_window}
In Table~\ref{tab:mAP_voc2007_test_sliding_windows} we provide the detection accuracy of our pipeline when, for generating the initial set of candidate boxes, we use a simple sliding windows scheme (of $10k$ windows per image).
We observe that:
\begin{itemize}
\item 
Even in this case, our pipeline achieves very high mAP results that are close to the ones obtained with selective search or edge box proposals. 
We emphasize that this is true even for the IoU$\geq0.7$ or the COCO style of mAP that favour better localized detections, despite the fact that in the case of sliding windows the initial set of candidate boxes is considerably less accurately localized than in the edge box or in the selective search cases (see Table~\ref{tab:recall_proposals}). 
\item 
In the case of sliding windows, just scoring the candidate boxes with the recognition model (hyphen (--) case) yields much worse mAP results than the selective search or the edge box proposals case. However, when we use the full detection pipeline that includes localization models and re-scoring of the new better localized candidate boxes, then this gap is significantly reduced. 
\item 
The difference in the mAP results between the proposed localization models (\emph{In-Out ML}, \emph{Borders ML}, and \emph{Combined ML}) and the \emph{bounding box regression} model (\emph{Bbox reg.}) is even greater in the case of sliding windows.
\end{itemize}

To the best of our knowledge, the above mAP results are considerably higher  than those of any other detection method when only sliding windows are used for  the initial bounding box proposals (similar experiments are reported in~\cite{girshick2015fast,hosang2015makes}).
We also note that we had not experimented with increasing the number of sliding windows. Furthermore,  the tested recognition model and localization models were not re-trained with sliding windows in the training set. 
As a result, we foresee that by exploring those two factors one might be able to further boost the detection performance for the sliding windows case. 

\subsection{Preliminary results on COCO} \label{sec:exps_coco}

To  obtain some preliminary results on COCO,  we applied our training procedure on COCO train set.  The only modification was to use $320k$ iterations (no other parameter was tuned). Therefore, LocNet results can still be significantly improved but the main goal was to show the relative difference in performance between the \emph{Combined ML} localization model and the box regression model. Results are shown in Table~\ref{tab:coco_minival}, where
it is observed that the proposed model boosts the mAP
by 5 points in the COCO-style evaluation, 8 points in the
$\mathrm{IoU}\geq0.75$ case and 1.4 points in the $\mathrm{IoU}\geq0.5$ case. 

\begin{table}
\centering
\renewcommand{\figurename}{Table}
\renewcommand{\captionlabelfont}{\bf}
\renewcommand{\captionfont}{\small} 
\captionsetup{labelsep=space}
\resizebox{0.48\textwidth}{!}{
{\setlength{\extrarowheight}{2pt}\scriptsize
{\begin{tabular}{l <{\hspace{-0.3em}}|>{\hspace{-0.5em}} l |>{\hspace{-0.5em}} l ||>{\hspace{-0.5em}} c | >{\hspace{-0.5em}}c | >{\hspace{-0.5em}}c | >{\hspace{-0.5em}}c }
\hline
\multicolumn{3}{c||}{Detection Pipeline} & \multicolumn{4}{c}{mAP }\\
\hline
Localization & Recognition & Proposals & Dataset & IoU $\geq$ 0.5 & IoU $\geq$ 0.75 & \emph{COCO} style\\
\hline
\emph{Combined ML} & \emph{Fast R-CNN} & \emph{Sel. Search} & 
$5K$ mini-val set & \textbf{0.424} & \textbf{0.282} & \textbf{0.264}\\
\emph{Bbox reg.}   & \emph{Fast R-CNN} & \emph{Sel. Search} & 
$5K$ mini-val set & 0.407 & 0.202 & 0.214\\
\hline
\emph{Combined ML} & \emph{Fast R-CNN} & \emph{Sel. Search} &
test-dev set & 0.429 & 0.279 & 0.263\\
\hline
\end{tabular}}}}
\vspace{1pt}
\caption{\small{\textbf{-- Preliminary results on COCO.} 
In those experiments the \emph{Fast R-CNN} recognition model uses
a softmax classifier~\cite{girshick2015fast} instead of class-specific linear SVMs~\cite{girshick2014rich} that are being used for the PASCAL experiments.}}
\label{tab:coco_minival}
\end{table}

\subsection{Qualitative results}  \label{sec:qualitative}
\begin{figure*}[t!]
\center
\renewcommand{\figurename}{Figure}
\renewcommand{\captionlabelfont}{\bf}
\renewcommand{\captionfont}{\small} 
\vspace{5pt}
\begin{subfigure}[b]{\textwidth}
\center
\renewcommand{\figurename}{Figure}
\renewcommand{\captionlabelfont}{\bf}
\renewcommand{\captionfont}{\footnotesize} 
        \begin{center}
        \begin{subfigure}[b]{0.16\textwidth}
        \includegraphics[width=\textwidth]{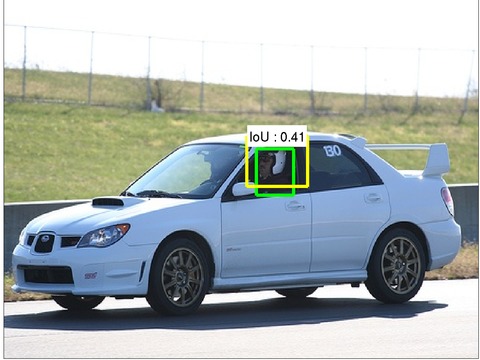}
        \end{subfigure} 
        \hspace{0.05cm} 
        \begin{subfigure}[b]{0.16\textwidth}
        \includegraphics[width=\textwidth]{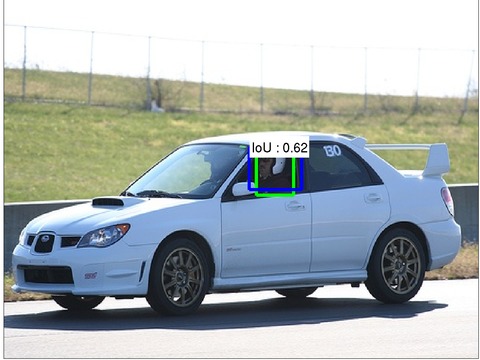}
        \end{subfigure} 
        \hspace{0.05cm} 
        \begin{subfigure}[b]{0.16\textwidth}
        \includegraphics[width=\textwidth]{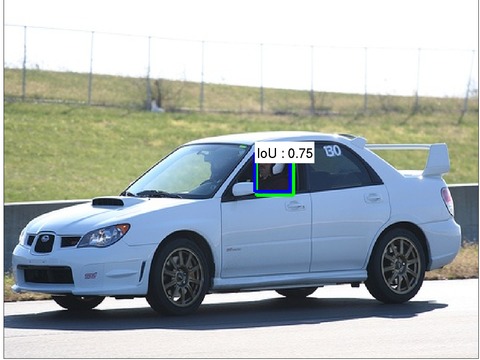}
        \end{subfigure} 
        \hspace{0.05cm} 
        \begin{subfigure}[b]{0.16\textwidth}
        \includegraphics[width=\textwidth]{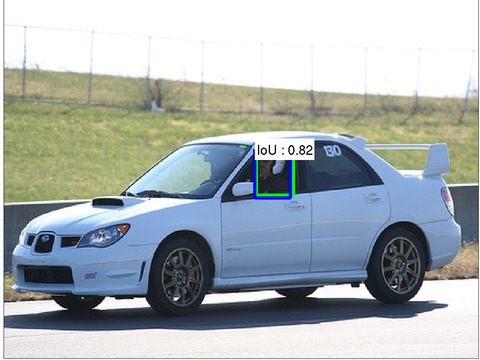}
        \end{subfigure} 
        \hspace{0.05cm} 
        \begin{subfigure}[b]{0.16\textwidth}
        \includegraphics[width=\textwidth]{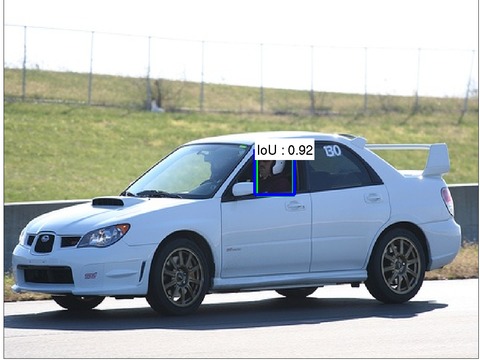}
        \end{subfigure} 
        \end{center}
        \vspace{5pt}
\end{subfigure}  
\begin{subfigure}[b]{\textwidth}
\center
\renewcommand{\figurename}{Figure}
\renewcommand{\captionlabelfont}{\bf}
\renewcommand{\captionfont}{\footnotesize} 
        \begin{center}
        \begin{subfigure}[b]{0.16\textwidth}
        \includegraphics[width=\textwidth]{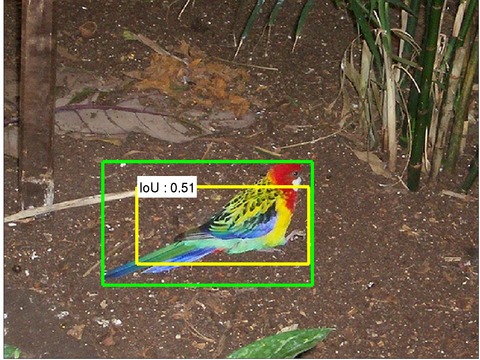}
        \end{subfigure} 
        \hspace{0.05cm} 
        \begin{subfigure}[b]{0.16\textwidth}
        \includegraphics[width=\textwidth]{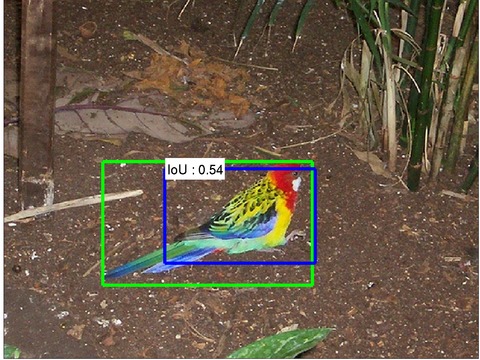}
        \end{subfigure} 
        \hspace{0.05cm} 
        \begin{subfigure}[b]{0.16\textwidth}
        \includegraphics[width=\textwidth]{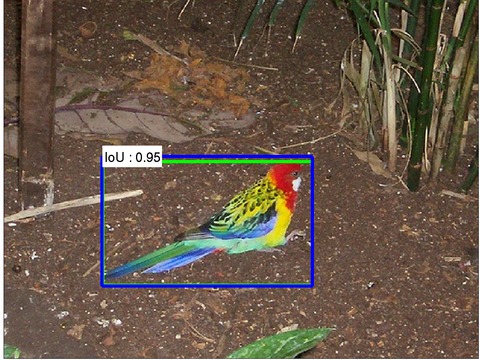}
        \end{subfigure} 
        \hspace{0.05cm} 
        \begin{subfigure}[b]{0.16\textwidth}
        \includegraphics[width=\textwidth]{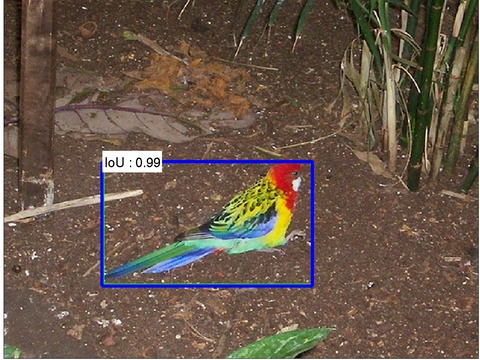}
        \end{subfigure} 
        \hspace{0.05cm} 
        \begin{subfigure}[b]{0.16\textwidth}
        \includegraphics[width=\textwidth]{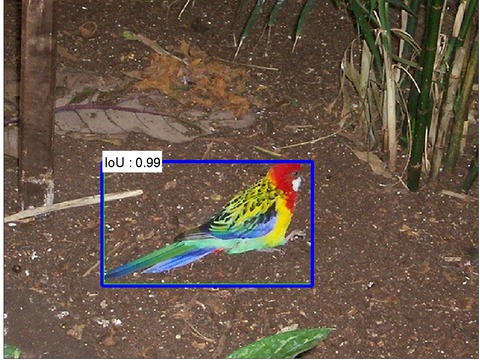}
        \end{subfigure} 
        \end{center}
        \vspace{5pt}
\end{subfigure} 
\begin{subfigure}[b]{\textwidth}
\center
\renewcommand{\figurename}{Figure}
\renewcommand{\captionlabelfont}{\bf}
\renewcommand{\captionfont}{\footnotesize} 
        \begin{center}
        \begin{subfigure}[b]{0.16\textwidth}
        \includegraphics[width=\textwidth]{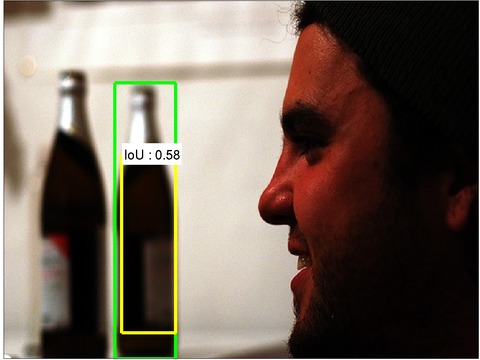}
        \end{subfigure} 
        \hspace{0.05cm} 
        \begin{subfigure}[b]{0.16\textwidth}
        \includegraphics[width=\textwidth]{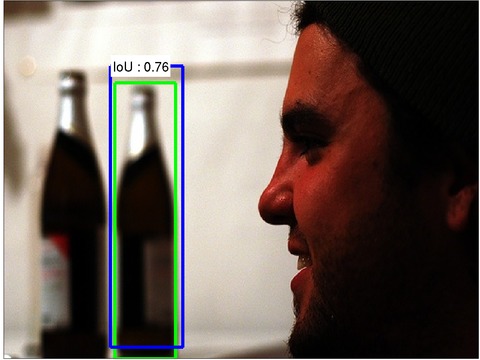}
        \end{subfigure} 
        \hspace{0.05cm} 
        \begin{subfigure}[b]{0.16\textwidth}
        \includegraphics[width=\textwidth]{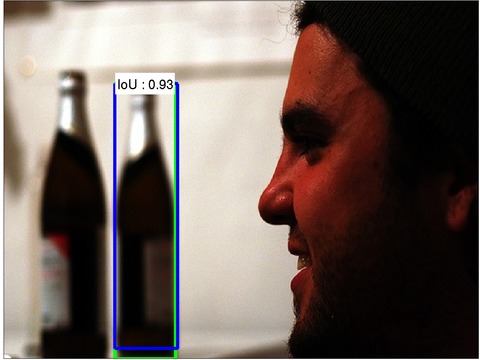}
        \end{subfigure} 
        \hspace{0.05cm} 
        \begin{subfigure}[b]{0.16\textwidth}
        \includegraphics[width=\textwidth]{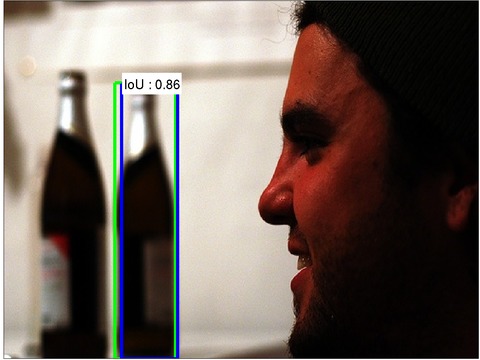}
        \end{subfigure} 
        \hspace{0.05cm} 
        \begin{subfigure}[b]{0.16\textwidth}
        \includegraphics[width=\textwidth]{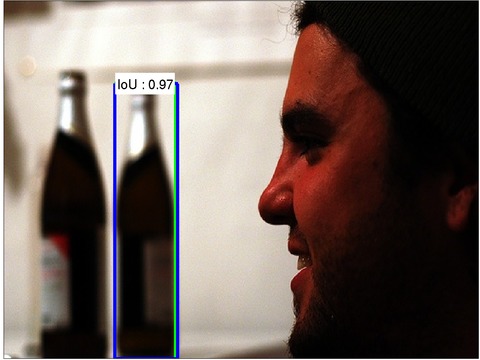}
        \end{subfigure} 
        \end{center}
        \vspace{5pt}
\end{subfigure}
\begin{subfigure}[b]{\textwidth}
\center
\renewcommand{\figurename}{Figure}
\renewcommand{\captionlabelfont}{\bf}
\renewcommand{\captionfont}{\footnotesize} 
        \begin{center}
        \begin{subfigure}[b]{0.16\textwidth}
        \includegraphics[width=\textwidth]{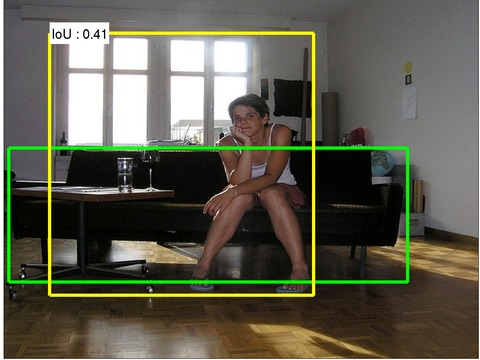}
        \end{subfigure} 
        \hspace{0.05cm} 
        \begin{subfigure}[b]{0.16\textwidth}
        \includegraphics[width=\textwidth]{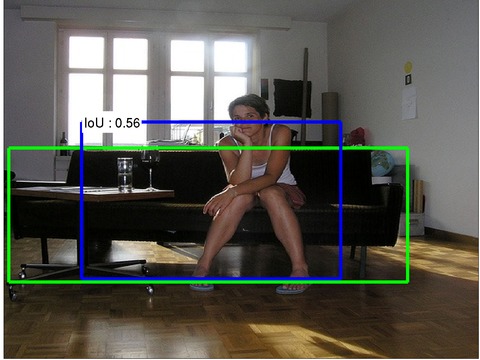}
        \end{subfigure} 
        \hspace{0.05cm} 
        \begin{subfigure}[b]{0.16\textwidth}
        \includegraphics[width=\textwidth]{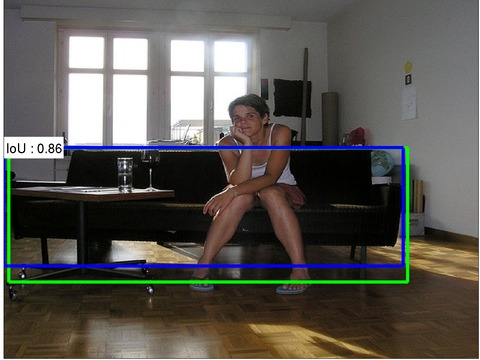}
        \end{subfigure} 
        \hspace{0.05cm} 
        \begin{subfigure}[b]{0.16\textwidth}
        \includegraphics[width=\textwidth]{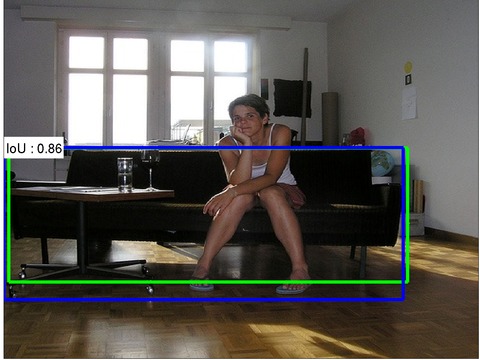}
        \end{subfigure} 
        \hspace{0.05cm} 
        \begin{subfigure}[b]{0.16\textwidth}
        \includegraphics[width=\textwidth]{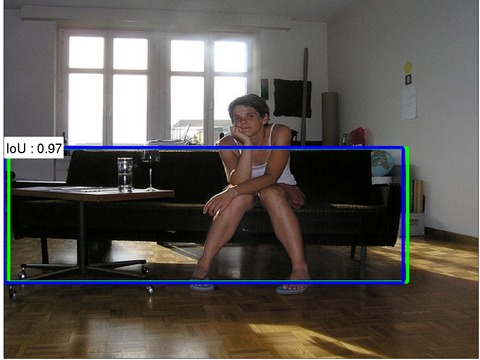}
        \end{subfigure} 
        \end{center}
        \vspace{5pt}
\end{subfigure}
\begin{subfigure}[b]{\textwidth}
\center
\renewcommand{\figurename}{Figure}
\renewcommand{\captionlabelfont}{\bf}
\renewcommand{\captionfont}{\footnotesize} 
        \begin{center}
        \begin{subfigure}[b]{0.16\textwidth}
        \includegraphics[width=\textwidth]{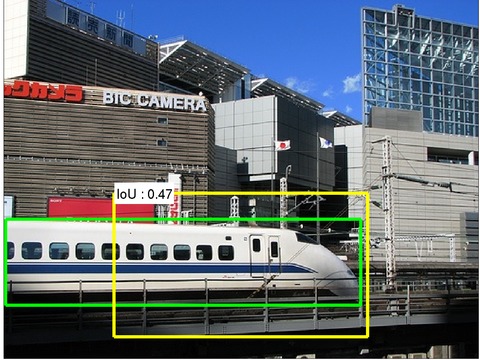}
        \end{subfigure} 
        \hspace{0.05cm} 
        \begin{subfigure}[b]{0.16\textwidth}
        \includegraphics[width=\textwidth]{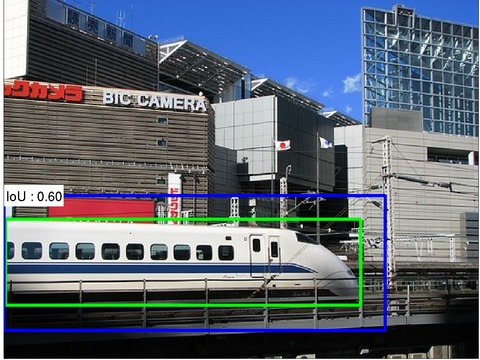}
        \end{subfigure} 
        \hspace{0.05cm} 
        \begin{subfigure}[b]{0.16\textwidth}
        \includegraphics[width=\textwidth]{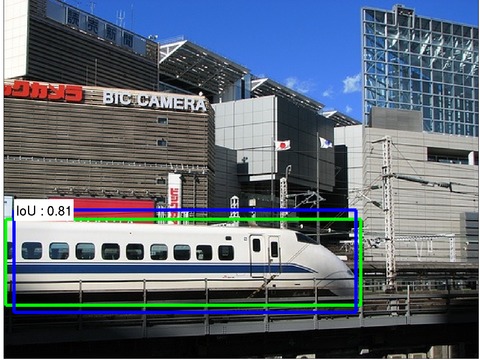}
        \end{subfigure} 
        \hspace{0.05cm} 
        \begin{subfigure}[b]{0.16\textwidth}
        \includegraphics[width=\textwidth]{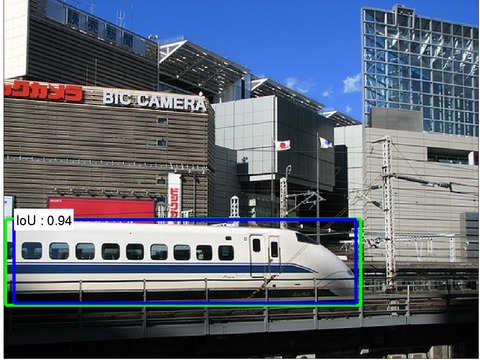}
        \end{subfigure} 
        \hspace{0.05cm} 
        \begin{subfigure}[b]{0.16\textwidth}
        \includegraphics[width=\textwidth]{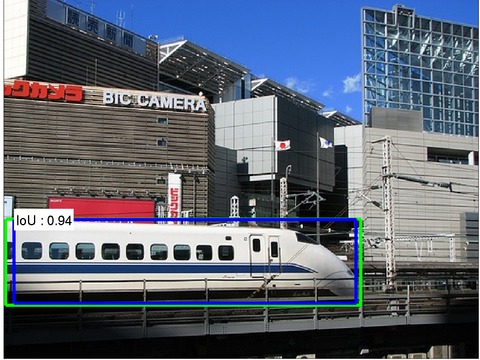}
        \end{subfigure} 
        \end{center}
        \vspace{5pt}
\end{subfigure}
\begin{subfigure}[b]{\textwidth}
\center
\renewcommand{\figurename}{Figure}
\renewcommand{\captionlabelfont}{\bf}
\renewcommand{\captionfont}{\footnotesize} 
        \begin{center}
        \begin{subfigure}[b]{0.16\textwidth}
        \includegraphics[width=\textwidth]{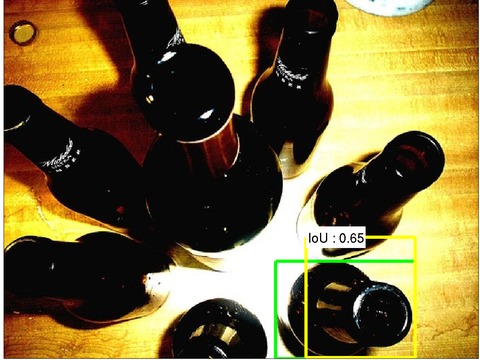}
        \end{subfigure} 
        \hspace{0.05cm} 
        \begin{subfigure}[b]{0.16\textwidth}
        \includegraphics[width=\textwidth]{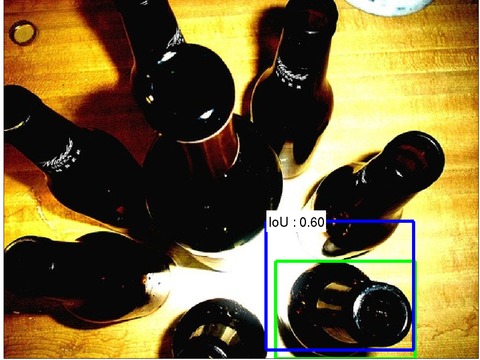}
        \end{subfigure} 
        \hspace{0.05cm} 
        \begin{subfigure}[b]{0.16\textwidth}
        \includegraphics[width=\textwidth]{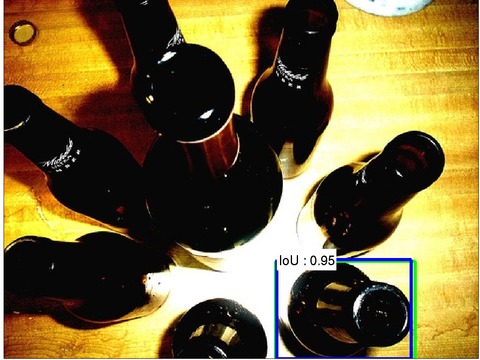}
        \end{subfigure} 
        \hspace{0.05cm} 
        \begin{subfigure}[b]{0.16\textwidth}
        \includegraphics[width=\textwidth]{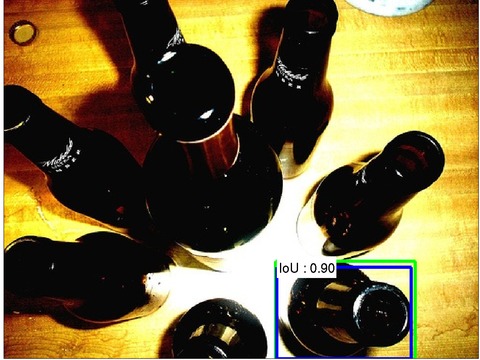}
        \end{subfigure} 
        \hspace{0.05cm} 
        \begin{subfigure}[b]{0.16\textwidth}
        \includegraphics[width=\textwidth]{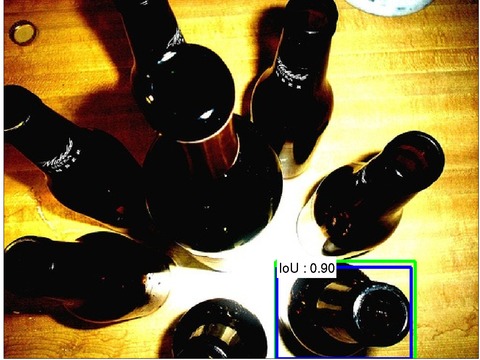}
        \end{subfigure} 
        \end{center}
        \vspace{5pt}
\end{subfigure}
\begin{subfigure}[b]{\textwidth}
\center
\renewcommand{\figurename}{Figure}
\renewcommand{\captionlabelfont}{\bf}
\renewcommand{\captionfont}{\footnotesize} 
        \begin{center}
        \begin{subfigure}[b]{0.16\textwidth}
        \includegraphics[width=\textwidth]{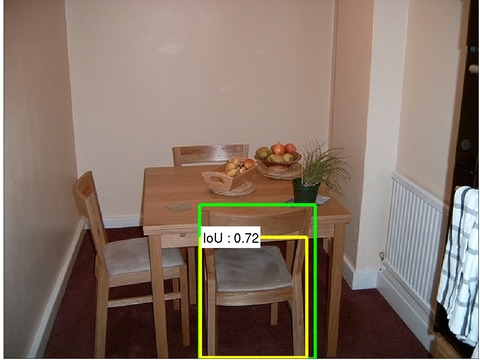}
        \caption{\small{Candidate box}}
        \end{subfigure} 
        \hspace{0.05cm} 
        \begin{subfigure}[b]{0.16\textwidth}
        \includegraphics[width=\textwidth]{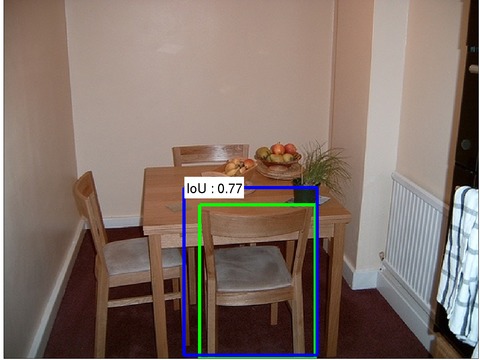}
        \caption{\small{Bbox regression}}
        \end{subfigure} 
        \hspace{0.05cm} 
        \begin{subfigure}[b]{0.16\textwidth}
        \includegraphics[width=\textwidth]{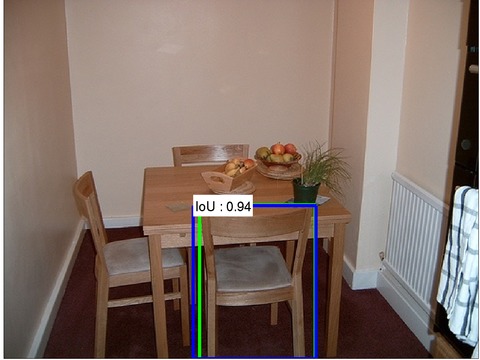}
        \caption{\small{In-Out ML}}
        \end{subfigure} 
        \hspace{0.05cm} 
        \begin{subfigure}[b]{0.16\textwidth}
        \includegraphics[width=\textwidth]{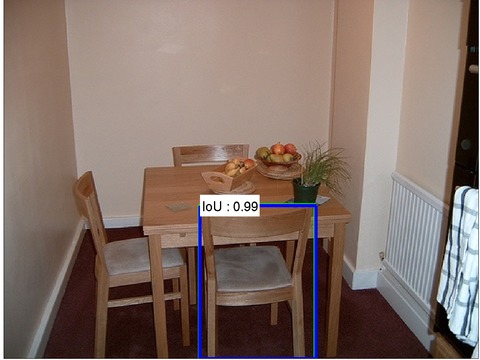}
        \caption{\small{Borders ML}}
        \end{subfigure} 
        \hspace{0.05cm} 
        \begin{subfigure}[b]{0.16\textwidth}
        \includegraphics[width=\textwidth]{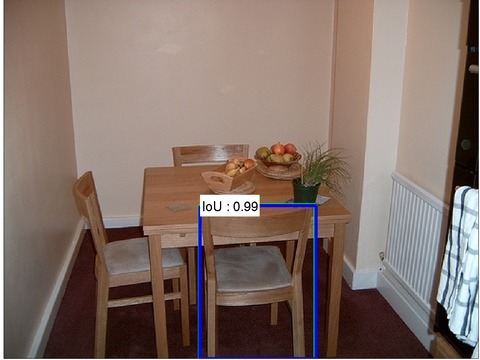}
        \caption{\small{Combined ML}}
        \end{subfigure} 
        \end{center}
        \vspace{5pt}
\end{subfigure}
\vspace{-7pt}
\caption{Qualitative results of the bounding box localization step given an initial candidate box (column \emph{(a)}) from the bounding box regression model (column \emph{(b)}), the \emph{In-Out ML} localization model (column \emph{(c)}), the \emph{Borders ML} localization model (column \emph{(d)}), and the \emph{Combined ML} localization model (column \emph{(e)}). The candidate box is drawn with yellow color, the predicted boxes are drawn with blue color, and the ground truth bounding box is drawn with green color.}
        \label{fig:Qualitative}
        \vspace{-5pt}

\end{figure*}

In Figure~\ref{fig:Qualitative} we provide sample qualitative results 
that compare the newly proposed localization models (\emph{In-Out ML}, \emph{Borders ML}, and \emph{Combined ML}) with the current state-of-the-art \emph{bounding box regression} localization model.

\section{Conclusion} \label{sec:conclusions}

We proposed a novel object localization methodology that is based on assigning probabilities related to the localization task on each row and column of the region in which it searches the object.
Those probabilities provide useful information regarding the location of the object inside the search region and they can be exploited in order to infer its boundaries with high accuracy. 
 
We implemented our model via using a convolutional neural network architecture  properly adapted for this task, called LocNet, and we extensively evaluated it on PASCAL VOC2007 test set. 
We demonstrate that it outperforms  CNN-based bounding box regression on all the evaluation metrics and it leads to a significant improvement on those metrics that reward good localization. 
Importantly, LocNet can be easily plugged into existing state-of-the-art object detection methods, in which case we show that it contributes to significantly boosting their performance.
Finally, we demonstrate that our object detection methodology can achieve very high mAP results even when 
the initial set of candidate boxes is generated by a simple sliding windows scheme.

{\small
\bibliographystyle{ieee}
\bibliography{my_paper_bib_short}
}

\appendix
\section{Bounding box regression model} \label{sec:bbox_reg}
This localization model consists of four CNN-based scalar regression functions $f_x(B,c)$, $f_y(B,c)$, $f_w(B,c)$, and $f_h(B,c)$ that 
given a candidate box $B$ and a category $c$, they actually predict the coefficients of a geometric transformation that will ideally map the candidate box $B$ to a ground truth bounding box of the $c$ object category~\cite{girshick2014rich}.
Specifically, if $B=(B_x,B_y,B_w,B_h)$ are the coordinates of the candidate box in form of its top-left corner $(B_x,B_y)$ and its width and height $(B_w,B_h)$, then the predicted candidate box $\hat{B}$ is given by the following equations:
\begin{align}
  \hat{B}_{x} & = B_w \cdot f_x(B,c) + B_x \\
  \hat{B}_{y} & = B_h \cdot f_y(B,c) + B_y \\
  \hat{B}_{w} & = B_w \cdot exp(f_w(B,c)) \\
  \hat{B}_{h} & = B_h \cdot exp(f_h(B,c)) \text{.}
\end{align} 
Hence, the four scalar target regression values $T=\{t_x,t_y,t_w,t_h\}$ for the ground truth bounding box $B^{gt}=(B_{x}^{gt},B_{y}^{gt},B_{w}^{gt},B_{h}^{gt})$ are defined as:
\begin{align}
  t_x & = \frac{B_{x}^{gt} - B_x}{B_w} & t_y & = \frac{B_{y}^{gt} - B_y}{B_h}\\
  t_w & = log(\frac{B_{w}^{gt}}{B_{w}}) & t_h & = log(\frac{B_{h}^{gt}}{B_{h}})) \text{.}
\end{align} 

For the CNN architecture that implements the bounding box regression model we adopt the one proposed in~\cite{gidaris2015object}. 
As a loss function we use the sum of euclidean distances between the target values and the predicted values of each training sample.
The final fully connected layer is initialized from a Gaussian distribution with standard deviation of $0.01$. 
The rest training details (\ie SGD, mini-batch, definition of training samples) are similar to those described for the proposed localization models.

\section{Training the localization models} \label{sec:train_loc_models}

As proposed in Fast-RCNN~\cite{girshick2015fast},
when training the \emph{bounding box regression} model we simultaneously train the Fast-RCNN recognition model with the two models sharing their convolutional layers.
In our experiments, this way of training improves the accuracy of both the Fast-RCNN recognition model and the bounding box regression model. 

On the contrary, the newly proposed localization models  (\ie, \emph{Borders ML}, \emph{In-Out ML}, and \emph{Combined ML}) are not currently  trained simultaneously with the recognition model.
We expect that the joint training of these models with the recognition model can help to further improve their overall performance, which is therefore something that we plan to explore in the future. 

\section{Region activation maps size in bounding box regression model} \label{sec:exps_size}
Early in our work we tried to boost the localization performance of the VGG16-based box regression model by increasing the size of the region activation maps from 7x7 to 14x14. 
The resulting network had around $422M$ parameters, and in order to train it successfully we had to initialize the fc6 layer by up-sampling with bilinear interpolation the initial 7x7x512x4096 weights (pre-trained on ImageNet) of it to 14x14x512x4096 and then divide them by a factor of $\frac{14^2}{7^2}$ (to leave unchanged the expected magnitude of its activations). 
Even with this initialization scheme, the difference w.r.t. the 7x7 case is negligible (see Table~\ref{tab:mAP_voc2007_test_roi_size}) and thus this models was not used.
Note that, despite the fact that the region activation maps of LocNet are of size  14x14x512, it contains only $26M$ and $34M$ parameters for the \emph{InOut} and \emph{Borders} cases correspondingly  (which is almost 15 times less than the 14x14 regression case) and it does not need any explicit care in the initialization of its extra layers.  

\begin{table}
\centering
\renewcommand{\figurename}{Table}
\renewcommand{\captionlabelfont}{\bf}
\renewcommand{\captionfont}{\small} 
\resizebox{0.45\textwidth}{!}{
{\setlength{\extrarowheight}{2pt}\scriptsize
{\begin{tabular}{l <{\hspace{-0.3em}}|>{\hspace{-0.5em}} c || >{\hspace{-0.5em}}c | >{\hspace{-0.5em}}c | >{\hspace{-0.5em}}c }
\hline
\multicolumn{2}{c||}{Localization model} & \multicolumn{3}{c}{mAP VOC2007 test}\\
\hline
Method & Region Feat. Size & IoU $\geq$ 0.5 & IoU $\geq$ 0.7 & \emph{COCO} style\\
\hline
\emph{Bbox reg.}             & 14 x 14 x 512 & 0.777 & 0.568 & 0.453\\
\emph{Bbox reg.}             & 7 x 7 x 512   & 0.777 & 0.570 & 0.452\\
\hline
\emph{InOut ML}              & 14 x 14 x 512 & 0.783 & \textbf{0.654} & 0.522\\
\emph{Borders ML}            & 14 x 14 x 512 & 0.780 & 0.644 & 0.525\\
\emph{Combined ML}           & 14 x 14 x 512 & \textbf{0.784} & 0.650 & \textbf{0.530}\\
\hline
\end{tabular}}}}
\vspace{1pt}
\caption{\small{mAP results on VOC2007 test set. All the entries use the \emph{Reduced-MR-CNN} recognition model and \emph{Edge Box} proposals for the initial set of boxes. 
In the second column (Region Feat. Size) we provide the region activation maps size that the region adaptive max-pooling layer yields.}}
\label{tab:mAP_voc2007_test_roi_size}
\vspace{-12pt}
\end{table}

\section{Recognition models} \label{sec:rec_models}

\emph{\textbf{Reduced MR-CNN model:}}  
We based the implementation of this model on the state-of-the-art MR-CNN detection system that was proposed in~\cite{gidaris2015object}.
Briefly, the MR-CNN detection system recognises  a candidate box by focusing on multiple regions of it, each with a dedicated network component termed as region adaptation module.
In our implementation however, for efficiency reasons and in order to speed up the experiments, we applied the following reductions:
\begin{itemize}
\item We include only six out of the ten regions proposed, by skipping the half regions.
\item We do not include the semantic segmentation-aware CNN features. 
\item We reduce the total amount parameters on the region adaptation modules.
\end{itemize}

In order to achieve the reduction of parameters on the hidden fully connected layers fc6 and fc7 of the region adaptation modules,
each of them is decomposed on two fully connected layers without any non-linearities between them. 
Specifically, the fc6 layer with weight matrix $W_6: 25088 \times 4096$ is decomposed on the layers fc6\_1 and fc6\_2 with weight matrices $W_{6.1}: 25088 \times 1024$ and $W_{6.2}: 1024 \times 4096$ correspondingly. The fc7 layer with weight matrix $W_7: 4096 \times 4096$ is decomposed on the layers fc7\_1 and fc7\_2 with weight matrices $W_{7.1}: 4096 \times 256$ and $W_{7.2}: 256 \times 4096$ correspondingly.
To train the Reduced MR-CNN network, we first train only the original candidate box region of it without reducing the parameters of the fc6 and fc7 layers. Then, we apply the truncated SVD decomposition on the aforementioned layers (for more details see section ~\S3.1 of ~\cite{girshick2015fast}) that results on the layers  fc6\_1, fc6\_2, fc7\_1, and fc7\_2. We copy the parameters of the resulting fully connected layers to the corresponding layers of the rest region adaptation modules of the model and we continue training. 

\emph{\textbf{Fast-RCNN model:}} 
We re-implemented Fast-RCNN based on the publicly available code provided from Fast-RCNN~\cite{girshick2015fast} and Faster-RCNN~\cite{shaoqing2015faster}. 
Here we will describe only the differences of our implementation with the original Fast-RCNN system~\cite{girshick2015fast}.
In our implementation, we have different branches for the recognition of a candidate box and for its bounding box regression
after the last convlutional layer (conv5\_3 for the VGG16-Net~\cite{simonyan2014very}) that do not share any weights.
In contrary, the original Fast-RCNN model splits to two branches after the last hidden layer.
We applied this modification because, in our case, the candidate box that is fed to the regression branch is enlarged by a factor $\alpha=1.3$ while the candidate box that is fed to recognition branch is not. 
Also, after the fine-tuning, we remove the softmax layer of the recognition branch and we train linear SVMs on top of the features that the last hidden layer of the recognition branch yields, just as R-CNN~\cite{girshick2014rich} does. 
Finally, we do not reduce the parameters of the fully connected layers by applying the truncated SVD decomposition on them as in the original paper.
In our experiments those changes improved the detection performance of the model.

\section{Object detection pipeline} \label{sec:pipeline_details}
In Algorithm~\ref{algo:detection} of section~\S\ref{sec:Methodology} we provide the pseudo-code of the object detection pipeline that we adopt.
For clarity purposes, the pseudo-code that is given corresponds to the single object category detection case and not the multiple object categories case that we are dealing with. 
Moreover, the actual detection algorithm, after scoring the candidate boxes for the first time  $t=1$, it prunes the candidate boxes with low confidence in order to reduce the computational burden of the subsequent iterations. 
For that purpose, we threshold the candidate boxes of each category such that their average number per image and per category to be around $18$ boxes. 
Also, during this step, non-max-suppression with IoU of $0.95$ is applied in order to remove near duplicate candidate boxes (in the case of using sliding windows to generate the initial set of candidate boxes this IoU threshold is set to $0.85$). A more detailed algorithm of our detection pipeline is presented in Algorithm~\ref{algo:detection2}. Note that, since the initial candidate boxes $\{\textbf{B}^1_c\}_{c=1}^{C}$ are coming from a category-agnostic bounding box proposal algorithm, those boxes are the same for all the categories and 
when applying on them (during $t=1$ iteration) the recognition module, the computation between all the categories can be shared. 

\begin{algorithm}[t!]\label{algo:detection2}
\SetKwInOut{Input}{Input}
\SetKwInOut{Output}{Output}
\setcounter{algocf}{1}
\Input{Image $\textbf{I}$, initial set of candidate boxes $\{\textbf{B}^1_c\}_{c=1}^{C}$} 
\Output{Final list of per category detections $\{ \textbf{Y}_c \}_{c=1}^{C}$}

\For{$t \gets 1$ \textbf{to} $T$} {
\For{$c \gets 1$ \textbf{to} $C$} {
$\textbf{S}^{t}_c \gets \textit{Recognition}(\textbf{B}^{t}_c| \textbf{I},c)$ \\
\If{$t == 1$} {
$\{\textbf{S}^{t}_c,\textbf{B}^{t}_c\} \gets PruneCandidateBoxes(\{\textbf{S}^{t}_c,\textbf{B}^{t}_c\})$
}
}
\If{$t < T$} {
\For{$c \gets 1$ \textbf{to} $C$} {
$\textbf{B}^{t+1}_c \gets \textit{Localization}(\textbf{B}^{t}_c| \textbf{I},c)$ \\
}
}
}
\For{$c \gets 1$ \textbf{to} $C$} {
$\textbf{D}_c \gets \cup_{t=1}^{T}{\{\textbf{S}^{t}_c,\textbf{B}^{t}_c\}}$ \\
$\textbf{Y}_c \gets \textit{PostProcess}(\textbf{D}_c)$\\  
}
\caption{Object detection pipeline}\label{algo:detection2}
\end{algorithm}

\end{document}